\documentclass[a4paper,fleqn]{cas-sc}

\usepackage[authoryear,longnamesfirst]{natbib}
\usepackage{booktabs}
\usepackage{longtable}
\usepackage{tabularx}
\usepackage{array}
\usepackage{graphicx}
\usepackage{algorithm}
\usepackage{algpseudocode}
\usepackage{float}

\usepackage{placeins}

\def\tsc#1{\csdef{#1}{\textsc{\lowercase{#1}}\xspace}}
\tsc{WGM}
\tsc{QE}
\tsc{EP}
\tsc{PMS}
\tsc{BEC}
\tsc{DE}

\begin{document}
\let\WriteBookmarks\relax
\def\floatpagepagefraction{1}
\def\textpagefraction{.001}



\title [mode = title]{Calibrated Tree--Neural Fusion for Fine-Grained Vegetation Community Classification}                  




\author[1]{Dristi Datta}
\cormark[1]
\ead{dristi.datta@sydneymet.edu.au}

\author[2]{Md Khalid Hasan Sakib}
\ead{khalidsakib121@gmail.com}

\author[3]{Manoranjan Paul}
\ead{mpaul@csu.edu.au}


\affiliation[1]{
    organization={School of Information Technology and Engineering,
    Sydney Metropolitan Institute of Technology (SydneyMet)},
    city={Sydney},
    state={NSW},
    postcode={2000},
    country={Australia}
}

\affiliation[2]{
    organization={Department of Electrical \& Electronic Engineering,
    Rajshahi University of Engineering \& Technology},
    city={Rajshahi},
    country={Bangladesh}
}

\affiliation[3]{
    organization={School of Computing, Mathematics and Engineering,
    Charles Sturt University},
    city={Bathurst},
    state={NSW},
    postcode={2795},
    country={Australia}
}

\cortext[cor1]{Corresponding author}

\begin{abstract}
Accurate vegetation-community classification is essential for ecological monitoring, habitat assessment, and evidence-based environmental management in heterogeneous landscapes. Existing studies often rely on standalone tree ensembles or generic neural networks, although fine-grained ecological classes frequently exhibit overlapping spectral, topographic, and structural characteristics. Many frameworks also provide limited protection against stacking leakage, insufficient probability calibration, weak minority-class evaluation, and little evidence of stability across repeated data splits. To address these limitations, this study proposes Calibrated EcoTreeFuseNet-Plus, a tree--neural probability-fusion framework that combines out-of-fold tree probabilities, EcoFuseNet-V2 outputs, validation-selected meta-learning, and post-hoc temperature scaling. Raster values from six LiDAR-derived terrain and canopy variables and two hyperspectral vegetation indices were extracted at coordinate-based reference locations. Quality control removed 26 samples with missing elevation and one sample with non-finite NDWI, producing 1,833 complete records across 29 vegetation and non-vegetation classes. On the held-out test set, the proposed model achieved an accuracy of 0.8000, a macro F1-score of 0.7768, a balanced accuracy of 0.7903, and an MCC of 0.7903. Calibration reduced the expected calibration error from 0.3866 to 0.0651 without changing class predictions. Five-seed evaluation yielded a macro F1-score of $0.7717 \pm 0.0112$, indicating stable performance across repeated splits. The results demonstrate a reliable discrimination--calibration trade-off for small-sample, fine-grained ecological classification.
\end{abstract}


\maketitle

\section{Introduction}
\label{sec:introduction}

Accurate vegetation-community classification is important for
biodiversity assessment, habitat monitoring, watershed management,
ecosystem restoration, and evidence-based conservation planning.
Vegetation communities reflect interacting gradients of species
composition, moisture availability, elevation, landform, solar
exposure, canopy structure, and disturbance history rather than
differences in surface cover alone. Reliable discrimination among
these communities is therefore valuable for characterizing ecological
organization and monitoring environmental change. The increasing
availability of airborne and satellite observations has expanded the
capacity to examine such patterns across spatially heterogeneous
landscapes
\citep{cavenderbares2022integrating,cavenderbares2025spectral,
foody2002status,stehman2019key}.

Fine-grained vegetation-community classification nevertheless remains
more difficult than conventional broad land-cover mapping.
Ecologically distinct communities may exhibit similar spectral
responses, whereas observations from the same community may vary
because of phenology, canopy density, water availability, terrain
illumination, substrate, and local environmental conditions. These
difficulties are particularly relevant in mountainous landscapes,
where vegetation distributions are strongly associated with
topographic and moisture gradients. A single predictor family may
therefore provide an incomplete representation of the ecological
setting. Spectral variables can characterize vegetation greenness and
moisture-related response, while LiDAR-derived products can provide
complementary information on terrain configuration and canopy
structure. Integrating these information domains consequently offers
a physically meaningful basis for fine-grained ecological
classification
\citep{talukdar2020landuse,cavenderbares2022integrating,
cavenderbares2025spectral,albanwan2024imagefusion}.

Remote-sensing vegetation analysis has long benefited from
physically interpretable spectral measurements and vegetation indices.
Normalized-difference formulations based on red, near-infrared, and
shortwave-infrared reflectance provide indicators related to
vegetation greenness, photosynthetic activity, and liquid-water
conditions
\citep{rouse1974monitoring,bannari1995review,gao1996ndwi}. More recent remote-sensing studies have further demonstrated that
vegetation interference, soil--vegetation spectral interactions, and
the design or transformation of spectral representations can
substantially influence environmental information retrieval from
multispectral and hyperspectral observations
\citep{datta2023dryindices,datta2024soilvegetation,
datta2026reflectgan}.
Although such indices remain ecologically useful, individual spectral
indicators or manually defined thresholds are insufficient when
numerous communities occupy overlapping regions of spectral feature
space. Fine-class discrimination therefore increasingly relies on
supervised learning methods capable of modelling nonlinear
relationships among multiple environmental predictors.

Tree-based ensemble methods have become particularly effective for
this type of structured classification problem. Random Forest combines
bootstrap aggregation and randomized feature selection
\citep{breiman2001random}, gradient boosting sequentially combines
weak learners to reduce predictive error
\citep{friedman2001greedy}, and Extremely Randomized Trees increases
ensemble diversity through additional randomization of candidate
splits \citep{geurts2006extremely}. These approaches are well suited
to heterogeneous tabular predictors because they can accommodate
nonlinear responses, interactions, non-Gaussian distributions, and
comparatively limited sample sizes. Their effectiveness in remote
sensing and environmental classification makes them important
reference models for evaluating more complex learning systems
\citep{rodriguez2012assessment,talukdar2020landuse,
venter2021continental}.

Subsequent boosting implementations further strengthened tabular
learning. XGBoost introduced scalable regularized gradient boosting
\citep{chen2016xgboost}, LightGBM improved computational efficiency
through histogram-based learning and leaf-wise tree growth
\citep{ke2017lightgbm}, and CatBoost introduced ordered boosting
procedures intended to reduce prediction shift
\citep{prokhorenkova2018catboost}. Together with Random Forest,
ExtraTrees, and histogram-based gradient boosting, these methods form
a diverse set of strong nonlinear classifiers. However, conventional
tree ensembles operate primarily through recursive partitioning of
the available predictors and do not naturally provide
modality-specific latent representations, adaptive feature-fusion
mechanisms, or prototype-based representation structures.

Neural models offer complementary representational capabilities.
Deep learning has enabled hierarchical feature learning from spectral,
spatial, temporal, and multimodal remote-sensing observations
\citep{ball2017comprehensive,li2019deep}. Recent environmental remote-sensing studies have also combined
attention mechanisms with probabilistic neural modelling to emphasize
informative spectral variables and quantify predictive uncertainty,
while transformer-guided learning has been investigated for detecting
and correcting noisy spectral observations
\citep{datta2025hbam,paul2025transformer}. In hyperspectral and
multitemporal classification, neural architectures have been used to
model complex spectral dependence and temporal or spatial context
\citep{jia2021survey,mazzia2020improvement}. Their advantages,
however, are most apparent when sufficiently large and representative
labelled datasets are available.

This requirement creates an important distinction for ecological
tabular classification. Many ecological studies contain limited
labelled observations, numerous imbalanced classes, and compact
predictor matrices derived from physically different environmental
sources. Under these conditions, neural superiority cannot be assumed.
Several architectures have attempted to improve neural learning for
tabular data, including Neural Oblivious Decision Ensembles
\citep{popov2020node}, TabNet \citep{arik2021tabnet}, and
TabTransformer \citep{huang2020tabtransformer}, while broader
comparisons have evaluated multilayer perceptrons, residual networks,
and transformer-based models for structured datasets
\citep{gorishniy2021revisiting}. Model-comparison studies using hyperspectral-derived environmental
predictors have likewise shown that predictive performance depends
strongly on both the learning algorithm and the representation of the
available spectral information
\citep{datta2023comparative}. Large empirical studies nevertheless
show that tree-based ensembles frequently remain highly competitive,
and often more stable, on small- and medium-scale tabular problems
\citep{shwartzziv2022tabular,grinsztajn2022why}. This evidence
suggests that, for limited ecological datasets, combining tree and
neural information may be more defensible than assuming that a
standalone neural architecture should replace strong tree learners.

Multimodal learning provides a broader conceptual basis for such
integration. Information from different sensing or feature domains
may be combined at the input, representation, decision, or hybrid
level \citep{baltrusaitis2019multimodal}. In remote sensing,
hyperspectral and LiDAR observations are especially complementary:
hyperspectral measurements characterize wavelength-dependent surface
response, whereas LiDAR provides elevation and structural information
that may help distinguish spectrally similar classes. Previous studies
have investigated attention-based hyperspectral--LiDAR fusion
\citep{feng2021linear}, LiDAR-guided cross-attention and autoencoder
strategies \citep{yang2024lidar,yang2024unsupervised}, and the effect
of spectral-band organization on LiDAR-assisted classification
\citep{yang2025hslinets}.

Related vegetation studies have also demonstrated the usefulness of
structural information. Airborne LiDAR and deep learning have been
applied to tree-species discrimination using three-dimensional canopy
geometry \citep{vermeer2023norwegian,lin2024pctrees}, while
multitemporal spectral methods have addressed species classification
under class imbalance \citep{mouret2024tree}. Sensor-agnostic
geometric representations have additionally been investigated for
transfer between mobile and airborne laser-scanning systems
\citep{korkeala2025normalview}. These advances demonstrate the value
of multisensor vegetation analysis, but most operate on image patches,
temporal sequences, or point-cloud representations. They do not
directly address the smaller tabular regime considered here, in which
LiDAR-derived terrain and canopy rasters and hyperspectral-derived
vegetation indices are sampled at labelled geographic reference
locations.

This distinction has methodological consequences. A coordinate-based
ecological table does not preserve the dense spatial neighbourhood of
an image patch or the three-dimensional geometry of a raw point cloud.
Instead, each observation is represented by a small number of
physically interpretable environmental variables. Tree ensembles can
model nonlinear thresholds and interactions among these variables,
whereas a compact neural branch can provide modality-specific
transformations, learned embeddings, and auxiliary representation
structure. A hybrid strategy can therefore exploit complementary
model behaviours, provided that the additional complexity is evaluated
against strong standalone tree baselines rather than assumed to be
beneficial.

Evaluation design is equally important. Ecological observations are
spatially structured, and geographically nearby samples may share
environmental characteristics. Unconstrained random splitting can
therefore provide optimistic estimates when training and test samples
are spatially close. Spatially structured resampling and blocked
cross-validation have consequently been recommended for ecological
and geospatial prediction
\citep{roberts2017cross,valavi2019blockcv,ploton2020spatial}.
However, strict geographic separation can be difficult in a
small-sample, 29-class problem because rare classes may become absent
from one or more partitions. A class-preserving, spatially informed
partition combined with explicit spatial-proximity diagnostics
provides a practical compromise, although it should not be interpreted
as a strict geographical extrapolation experiment.

Stacking introduces a separate source of possible optimism. If a
meta-classifier is trained using branch predictions generated by
models fitted directly on those same observations, the resulting
meta-features contain information unavailable for unseen cases.
Stacked generalization therefore relies on held-out predictions for
higher-level learning \citep{wolpert1992stacked}, while the Super
Learner framework formalized cross-validated ensemble construction
\citep{vanderlaan2007super}. Out-of-fold prediction is consequently
important for defensible tree--neural fusion. A separate validation
partition is also useful for selecting candidate meta-classifiers,
soft-voting strategies, checkpoints, and calibration parameters
without directly optimizing those choices against the final test
labels.

Probability reliability constitutes another important but often
distinct aspect of classifier evaluation. Accuracy and macro F1
describe discrimination, but they do not indicate whether predictions
assigned a confidence of 0.90 are correct approximately 90\% of the
time. This distinction matters in ecological applications because
confidence estimates may be used to prioritize field verification or
identify predictions requiring additional scrutiny. Supervised
classifiers can differ substantially in the quality of their
probability estimates \citep{niculescu2005predicting}, and modern
learning systems may produce systematically miscalibrated confidence
scores \citep{guo2017calibration}. Probability calibration should
therefore be assessed alongside discrimination using measures such as
expected calibration error, Brier score, negative log-likelihood, and
reliability diagrams.

Taken together, these developments identify a specific methodological
gap. Fine-grained ecological classification from compact multisource
tables lies between conventional tabular machine learning and
high-dimensional multimodal deep learning. Strong tree ensembles
provide competitive discrimination, whereas neural models can offer
complementary probability and representation information; however,
the usefulness of combining these sources must be established under
a leakage-aware fusion protocol and evaluated against strong tree
baselines. At the same time, spatially informed partitioning,
class-sensitive evaluation, probability calibration, statistical
comparison, and repeated-split stability are required to avoid
drawing conclusions from a favourable partition or from hard-label
performance alone.

Accordingly, this study proposes \textit{Calibrated
EcoTreeFuseNet-Plus} for fine-grained vegetation-community
classification using LiDAR-derived terrain and canopy variables and
hyperspectral-derived vegetation indices. Coordinate-based raster
sampling and systematic quality control produced 1,833 complete
observations representing 29 vegetation and non-vegetation classes.
The eight predictors comprised elevation, slope, curvature,
Topographic Position Index, solar insolation, canopy height, NDVI, and
NDWI. Geographic coordinates and reference-source information were
retained for spatial and source diagnostics but were excluded from the
predictor matrix.

The framework combines six tree-probability branches---Random Forest,
XGBoost, ExtraTrees, HistGradientBoosting, LightGBM, and CatBoost---with
EcoFuseNet-V2, a compact neural branch that separately processes
terrain, canopy, and spectral predictors. Leakage-aware out-of-fold
generation provides tree probabilities and neural outputs for
meta-learning. The complete meta-representation contains the
standardized original predictors, six tree-probability vectors,
EcoFuseNet-V2 class probabilities, its ecological embedding, gate
representation, and prototype-distance features. Candidate
meta-classifiers and a validation-weighted soft-voting alternative are
evaluated using the predefined validation partition, after which the
selected raw probability source is calibrated using temperature
scaling. The resulting model is evaluated on an untouched test
partition using class-balanced discrimination metrics,
probability-reliability measures, class-wise analysis, paired
statistical comparisons, architectural and modality ablations, neural
diagnostics, and repeated-seed/split evaluation.

The principal contributions of this study are as follows:

\begin{itemize}

\item A quality-controlled ecological modelling dataset is constructed
from six LiDAR-derived terrain and canopy predictors and two
hyperspectral-derived vegetation indices sampled at labelled reference
locations, yielding 1,833 complete observations across 29 fine
ecological classes.

\item EcoFuseNet-V2 is developed as a compact modality-aware neural
branch that separately processes terrain, canopy structure, and
spectral information before gated fusion, ecological embedding, and
prototype-based auxiliary representation learning.

\item EcoTreeFuseNet-Plus is formulated as a leakage-aware hybrid
probability-fusion framework that combines six strong tree ensembles
with EcoFuseNet-V2 outputs through out-of-fold meta-feature
construction, validation-selected meta-learning, and a competing
validation-weighted soft-voting strategy.

\item Post-hoc temperature scaling is incorporated into the principal
model so that probability reliability is evaluated jointly with
predictive discrimination rather than treating hard-label accuracy as
the sole performance criterion.

\item The experimental protocol combines spatially informed
class-preserving partitioning, training-derived preprocessing,
class-sensitive learning, untouched test evaluation, bootstrap
confidence intervals, paired statistical comparisons, class-wise
analysis, ablation experiments, and neural-branch diagnostic analyses.

\item Five independently repeated seed/split experiments are used to
assess sensitivity to data partitioning and stochastic training, while
the final interpretation explicitly distinguishes calibration benefits
from discrimination gains relative to the strongest standalone tree
baseline.

\end{itemize}

The remainder of this manuscript is organized as follows.
Section~\ref{sec:study-area-dataset} introduces the East River study
area, the NEON Airborne Observation Platform dataset, the reference
observations, the target-class structure, and the final predictor set.
Section~\ref{sec:data-preparation} describes geospatial data
organization, coordinate-based raster extraction, quality control, and
construction of the final modelling table.
Section~\ref{sec:experimental-design} presents the research design,
feature transformations, spatially informed stratified partitioning,
class weighting, evaluation metrics, statistical procedures, and
reproducibility controls.
Section~\ref{sec:proposed-framework} describes Calibrated
EcoTreeFuseNet-Plus, including EcoFuseNet-V2, the tree-probability
branches, leakage-aware out-of-fold feature generation, hybrid
meta-learning, validation-weighted soft voting, calibration, and the
secondary model variants.
Section~\ref{sec:experimental-results} reports the experimental
results, including baseline comparisons, calibration performance,
statistical comparisons, ablation analyses, class-wise behaviour,
meta-feature and neural diagnostics, and repeated-split stability.
Section~\ref{sec:discussion} interprets the findings in relation to
fine-grained ecological classification, tabular learning, multisource
fusion, probability reliability, model interpretation, uncertainty,
and computational considerations.
Finally, Section~\ref{sec:conclusions} summarizes the principal
findings, remaining limitations, future research directions, and
practical relevance of the proposed framework.

\section{Study Area and Dataset}
\label{sec:study-area-dataset}

\subsection{Study Area and Airborne Observations}
\label{subsec:study-area}

The study was conducted in the East River watershed near Crested Butte, Colorado, USA. This mountainous headwater catchment forms part of the Upper Colorado River system and is characterized by pronounced topographic relief, complex geological conditions, seasonally snow-dominated hydrology, and strong spatial gradients in elevation, solar exposure, soil moisture, and vegetation structure \cite{hubbard2018eastRiver}. These interacting environmental controls support a heterogeneous landscape composed of alpine and subalpine communities, wet and dry meadows, herb-, shrub-, and grass-dominated vegetation, exposed geological surfaces, water bodies, snow-covered areas, and developed surfaces.

The environmental complexity of the watershed creates a demanding setting for fine-grained ecological classification. Communities with similar spectral characteristics may occur under contrasting topographic, geological, or moisture conditions, whereas observations belonging to the same community may vary because of phenology, canopy density, illumination, and local environmental heterogeneity. Consequently, no single information domain can be expected to characterize all ecological classes adequately.

The remote-sensing observations were acquired by the National Ecological Observatory Network Airborne Observation Platform during a survey conducted in June 2018 \cite{falco2023eastRiverDataset}. The NEON airborne system integrates hyperspectral imaging, LiDAR, and high-resolution photography to provide spatially coordinated measurements of vegetation properties, terrain, and landscape structure \cite{kampe2010neon,musinsky2022spanning}. Hyperspectral observations characterize surface-reflectance variations associated with vegetation condition, while LiDAR provides complementary information on terrain elevation and three-dimensional vegetation structure \cite{lefsky2002lidar}. This complementarity provides the observational basis for the multisource classification framework developed in this study.

\subsection{Data Source and Reference Observations}
\label{subsec:data-source}

The study used the publicly available East River vegetation-classification and environmental-covariate dataset distributed through the Environmental Systems Science Data Infrastructure for a Virtual Ecosystem repository \cite{falco2023eastRiverDataset}. The source package integrates labelled ecological observations with spatially aligned vegetation maps, LiDAR-derived terrain and canopy products, hyperspectral-derived vegetation indices, quality-control masks, and broader environmental covariates.

Only the coordinate-based reference observations and selected terrain, canopy, and spectral products were used to construct the predictive dataset. The vegetation-classification map and ancillary masks were retained as contextual products rather than model inputs. This separation avoided circular use of an existing classification product as a predictor of its associated ecological labels.

The reference information included sample identifiers, geographic and projected coordinates, fine-class labels, numerical class codes, subclass information, broader ecological types, and reference-source metadata. Projected coordinates in WGS 84 / UTM Zone 13N were used for spatial sampling and subsequent proximity diagnostics. Geographic and projected coordinates were excluded from the predictor matrix to prevent the models from learning direct location-based shortcuts.

The source reference set contained 1,860 observations. After coordinate-based extraction and quality screening, 1,833 complete observations were retained for modelling. The screening procedure and exclusion criteria are described in Section~\ref{sec:data-preparation}.

\subsection{Target-Class Structure}
\label{subsec:reference-targets}

The primary prediction task comprised 29 vegetation and non-vegetation classes derived from the source classification system \cite{falco2023eastRiverDataset}. The target structure included alpine and subalpine communities, wetland and meadow communities, herbaceous, shrub, and grass classes, exposed geological materials, water, snow, and developed or transportation surfaces.

The class distribution was uneven, with individual class sizes ranging from 16 to 178 observations. The largest-to-smallest class ratio was therefore

\begin{equation}
\frac{178}{16}=11.125.
\label{eq:class-imbalance-ratio}
\end{equation}

This imbalance motivated the use of stratified partitioning, class-sensitive training, macro-averaged evaluation, balanced accuracy, and class-wise performance analysis.

Each fine class was additionally associated with one of five broader ecological categories: \textit{Community}, \textit{Non-vegetated}, \textit{Herbs}, \textit{Shrubs}, and \textit{Grasses}. The 29 fine classes defined the principal training and evaluation objective. The coarse categories were retained only for auxiliary coarse-consistency learning and secondary hierarchy-aware analysis; they did not replace or modify the fine-class target.

\subsection{Predictor Variables}
\label{subsec:final-predictors}

Eight predictors were selected to represent complementary terrain, canopy-structure, and spectral characteristics. The terrain domain comprised elevation, slope, curvature, Topographic Position Index, and solar insolation. Canopy structure was represented by the Canopy Height Model, while spectral vegetation response was characterized using the Normalized Difference Vegetation Index and the Normalized Difference Water Index.

Terrain attributes provide information on landform configuration and environmental gradients that influence water redistribution, energy availability, soil development, and vegetation organization \cite{moore1991digital}. Elevation represents broad climatic and ecological zonation, while slope, curvature, and relative topographic position describe local terrain conditions. Solar insolation characterizes potential energy exposure. The Canopy Height Model contributes a structural dimension that cannot be represented directly by spectral vegetation indices \cite{lefsky2002lidar}.

NDVI summarizes vegetation greenness and photosynthetic activity through red and near-infrared reflectance, whereas NDWI represents vegetation liquid-water and moisture-related spectral responses \cite{rouse1974monitoring,gao1996ndwi}. Their combination with terrain and canopy measurements was intended to provide a more complete ecological representation than either modality alone.

The final predictor vector for observation $i$ was defined as

\begin{equation}
\mathbf{x}_{i}
=
\left[
x_{i}^{\mathrm{elev}},
x_{i}^{\mathrm{slope}},
x_{i}^{\mathrm{curv}},
x_{i}^{\mathrm{tpi}},
x_{i}^{\mathrm{solar}},
x_{i}^{\mathrm{chm}},
x_{i}^{\mathrm{ndvi}},
x_{i}^{\mathrm{ndwi}}
\right]^{\mathsf{T}}
\in \mathbb{R}^{8}.
\label{eq:eight-feature-vector}
\end{equation}

Table~\ref{tab:predictor-domains} summarizes the three information domains and their roles in the classification framework.

\begin{table}[htbp]
\centering
\caption{Predictor domains used in the vegetation-community classification framework.}
\label{tab:predictor-domains}
\begin{tabularx}{\textwidth}{
>{\raggedright\arraybackslash}p{0.20\textwidth}
>{\raggedright\arraybackslash}p{0.36\textwidth}
>{\raggedright\arraybackslash}X}
\toprule
\textbf{Information domain} &
\textbf{Predictors} &
\textbf{Principal ecological information} \\
\midrule

Terrain &
Elevation, slope, curvature, Topographic Position Index, and solar insolation &
Elevational gradients, terrain configuration, relative landscape position, water redistribution, and potential energy exposure \\

Canopy structure &
Canopy Height Model &
Vegetation structural height and differentiation among low-growing, taller, and non-vegetated surfaces \\

Spectral response &
Normalized Difference Vegetation Index and Normalized Difference Water Index &
Vegetation greenness, photosynthetic activity, and moisture-related spectral response \\
\bottomrule
\end{tabularx}
\end{table}

For modality-specific experiments, the six LiDAR-derived terrain and canopy variables formed the structural--topographic feature group, the two vegetation indices formed the spectral feature group, and all eight predictors formed the fused feature group. These definitions were maintained consistently across the baseline models, EcoFuseNet-V2, modality ablations, and the final tree--neural fusion framework.

\subsection{Dataset Quality and Analytical Scope}
\label{subsec:dataset-integrity}

All 1,833 retained observations contained finite values for the eight predictors. Missing or invalid predictor values were not replaced using synthetic, interpolated, or model-generated estimates. The complete coordinate-based extraction, screening, and verification procedure is described in Section~\ref{sec:data-preparation}.

The analytical unit of the study is a labelled geographical reference observation rather than an image patch, point-cloud segment, or complete raster scene. The task is therefore formulated as fine-grained ecological tabular classification. The combination of a limited sample size, 29 target classes, uneven class frequencies, overlapping ecological characteristics, and spatial dependence provides a challenging test case for evaluating calibrated tree--neural fusion under realistic small-sample conditions.

\section{Geospatial Data Preparation and Feature Extraction}
\label{sec:data-preparation}

\subsection{Extraction Workflow and Source Selection}
\label{subsec:extraction-workflow}

A coordinate-based workflow was implemented to transform the distributed airborne products into a machine-learning-ready tabular dataset. The principal steps comprised source-data organization, selection of spatially referenced ecological observations, verification of the predictor rasters, point-wise feature extraction, and systematic quality control. The source products were obtained from the East River vegetation-classification and environmental-covariate dataset described in Section~\ref{sec:study-area-dataset} \cite{falco2023eastRiverDataset}.

The downloaded materials were organized into classification-reference data, LiDAR-derived terrain and canopy products, hyperspectral-derived vegetation indices, classification maps, spatial masks, and metadata. Only the coordinate-based reference observations and eight predictor rasters were required for model development; the remaining products were retained for contextual verification.

The initially reviewed vegetation-reference table contained thematic class information but no geographic or projected coordinates. It could therefore describe the classification system but could not support direct point-to-raster extraction. A coordinate-complete reference table containing unique sample identifiers, fine-class labels, longitude, latitude, projected Easting and Northing, class metadata, coarse ecological types, and reference-source information was consequently selected.

Eight predictors were extracted: elevation, slope, curvature, Topographic Position Index, solar insolation, Canopy Height Model, NDVI, and NDWI. The first six represented LiDAR-derived terrain or structural information, while the final two represented hyperspectral-derived vegetation response. 

\subsection{Spatial Verification and Raster Sampling}
\label{subsec:raster-sampling}

The reference observations and selected raster products were represented in WGS 84 / UTM Zone 13N (EPSG:32613). This projected coordinate reference system expresses Easting and Northing in metres and is appropriate for coordinate-based analysis within the study region \cite{epsg32613}. Projected coordinates were therefore used for raster sampling, whereas longitude and latitude were retained only as reference metadata.

Before extraction, the coordinate reference system, spatial resolution, raster bounds, affine transformation, data type, and declared no-data metadata of each layer were inspected \cite{gdalRasterModel}. Because each raster was sampled independently at the same real-world locations, identical raster dimensions or grid origins were not required. Instead, the procedure required compatible coordinate systems and valid spatial coverage.

For observation $i$ and predictor $j$, the extracted value was defined as

\begin{equation}
x_{ij}
=
R_{j}\!\left(E_i,N_i\right),
\qquad
j=1,\ldots,8,
\label{eq:raster-point-sampling}
\end{equation}

where $R_j$ denotes the georeferenced raster and $(E_i,N_i)$ denotes the projected coordinate of the reference observation. The raster cell containing each coordinate was sampled using the native affine transformation of the corresponding layer \cite{rasterioSample}. No additional spatial interpolation or neighbourhood averaging was introduced during extraction.

Declared no-data values, masked cells, and locations outside valid raster coverage were treated as missing observations. Numerical zeros were retained as valid source values. The initial extraction produced 1,860 observations, with explicit missingness and finite-value indicators generated for subsequent screening.

\subsection{Quality Control and Final Dataset}
\label{subsec:extraction-quality-control}

Quality control was applied sequentially to the eight extracted
predictors. Of the initial 1,860 reference observations, 26 contained
missing elevation values and were removed, reducing the dataset to
1,834 observations. A subsequent finite-value screening identified one
additional observation with a negative-infinite NDWI value. Removal of
this observation yielded the final modelling sample:

\begin{equation}
1860 - 26 - 1 = 1833.
\label{eq:final-sample-count}
\end{equation}

No missing or non-finite predictor values were imputed, interpolated,
synthetically generated, or replaced through model-based estimation.
The resulting modelling table contained 1,833 observations and 21
analytical fields, comprising:

\begin{itemize}
    \item a unique sample identifier and the fine-class target;
    \item projected spatial coordinates and coordinate-reference
    information;
    \item eight terrain, canopy-structure, and spectral predictors;
    \item two quality-control indicators; and
    \item auxiliary reference metadata describing geographic
    coordinates, numerical class information, subclass, coarse
    ecological type, and reference source.
\end{itemize}

Only the eight environmental predictors were used as model inputs.
Projected and geographic coordinates, quality-control fields, coarse
ecological types, subclass information, and reference-source metadata
were retained exclusively for traceability, spatial diagnostics, and
post hoc descriptive analyses. The final predictor matrix contained no
missing, positive-infinite, or negative-infinite values. No duplicated
sample identifiers, projected-coordinate pairs, or complete
observations were detected. Blank subclass entries were retained as
valid auxiliary metadata because subclass information was not used for
model fitting or prediction.

\subsection{Class and Reference Distributions}
\label{subsec:data-distributions}

The final dataset comprised 29 fine vegetation and non-vegetation classes. Class frequencies ranged from 16 observations for BareSoil to 178 observations for IrrigLawnPasture, yielding a largest-to-smallest ratio of 11.125. This imbalance motivated the stratified, class-sensitive, and macro-averaged evaluation procedures introduced in Section~\ref{sec:experimental-design}.

\begin{table}[htbp]
\centering
\caption{Distribution of the 29 fine target classes after quality control.}
\label{tab:class-distribution}
\small
\begin{tabular}{lrlr}
\toprule
\textbf{Class} & \textbf{$n$} &
\textbf{Class} & \textbf{$n$} \\
\midrule
IrrigLawnPasture   & 178 & WetMeadowAlluv     & 50 \\
DrySubalpShale     & 166 & DrySubalpSandstone & 42 \\
AlpineSandstone    & 127 & AspenFringe        & 40 \\
AlpineGranite      & 109 & GraniteTalus       & 40 \\
DrySubalpTill      & 97  & StreamSandGravel   & 37 \\
AlpineShale        & 91  & GravelRoadTrail    & 33 \\
Artemisia          & 77  & ShaleBarren        & 32 \\
WetMeadowTill      & 73  & Paved              & 30 \\
EmergentWetland    & 73  & Wyethia            & 29 \\
DrySubalpGranite   & 72  & SandstoneOutcrop   & 26 \\
Festuca            & 70  & Snow               & 24 \\
GraniteOutcrop     & 70  & Frasera            & 18 \\
WetAlpine          & 70  & SandstoneTalus     & 18 \\
Water              & 66  & BareSoil           & 16 \\
Lupinus            & 59  &                    &    \\
\bottomrule
\end{tabular}
\normalsize
\end{table}

The coarse ecological-type and reference-source distributions are summarized in Table~\ref{tab:reference-distributions}. These variables were retained for descriptive and post hoc diagnostic analyses but were excluded from the predictor matrix.

\begin{table}[htbp]
\centering
\caption{Coarse ecological-type and reference-source distributions.}
\label{tab:reference-distributions}
\begin{tabular}{llrr}
\toprule
\textbf{Grouping} &
\textbf{Category} &
\textbf{$n$} &
\textbf{Percentage (\%)} \\
\midrule
Coarse type & Community      & 1,188 & 64.81 \\
            & Non-vegetated &   392 & 21.39 \\
            & Herbs          &   106 &  5.78 \\
            & Shrubs         &    77 &  4.20 \\
            & Grasses        &    70 &  3.82 \\
\midrule
Reference source & Supplemental & 1,580 & 86.20 \\
                 & Digitized    &   253 & 13.80 \\
\bottomrule
\end{tabular}
\end{table}

\subsection{Exploratory Predictor Analysis}
\label{subsec:exploratory-predictor-analysis}

Descriptive statistics were calculated before the model-specific transformations described in Section~\ref{sec:experimental-design}. The predictors differed substantially in scale and distribution, supporting the later use of training-set-based preprocessing for neural and meta-learning components.

\begin{table}[htbp]
\centering
\caption{Descriptive statistics of the eight retained predictors in their native numerical units.}
\label{tab:predictor-summary-statistics}
\small
\begin{tabular}{lrrrr}
\toprule
\textbf{Predictor} &
\textbf{Mean} &
\textbf{SD} &
\textbf{Minimum} &
\textbf{Maximum} \\
\midrule
Elevation          & 3,093.6127 & 353.6971    & 2,684.3799   & 3,999.4800 \\
Slope              & 14.3934    & 12.0482     & 0.0000       & 62.7826 \\
Curvature          & 0.0010     & 0.0159      & $-0.1028$    & 0.0776 \\
TPI                & 0.0048     & 0.0719      & $-0.5376$    & 0.8853 \\
Solar insolation   & 865,114.5820 & 270,289.8564 & 144,270.6563 & 1,249,377.2500 \\
CHM                & 0.0385     & 0.6052      & 0.0000       & 16.8799 \\
NDVI               & 0.4297     & 0.2870      & $-1.1194$    & 0.8660 \\
NDWI               & $-0.0660$  & 0.1899      & $-3.2500$    & 4.7143 \\
\bottomrule
\end{tabular}
\normalsize
\end{table}

The finite NDVI and NDWI ranges extended beyond the interval commonly associated with their standard normalized-difference formulations \cite{rouse1974monitoring,gao1996ndwi}. These values were retained because they originated from the distributed products and passed the finite-value checks. Arbitrary clipping was avoided because it would have introduced an additional transformation without an independently defined quality threshold.

The Canopy Height Model was strongly zero-inflated: 1,744 observations had a value of zero, corresponding to 95.14\% of the final dataset, while 89 observations contained positive values. These zeros were retained as valid source measurements. The pronounced sparsity and right-skewness motivated the logarithmic transformation and restrained canopy contribution described in the EcoFuseNet-V2 methodology.

Pairwise Pearson correlations were also calculated using the untransformed predictor values. The strongest positive relationship occurred between elevation and slope ($r=0.627$). Slope was negatively related to solar insolation ($r=-0.488$), elevation was moderately negatively related to solar insolation ($r=-0.390$), and curvature was moderately positively related to TPI ($r=0.378$). Other associations were weaker, including elevation--NDVI ($r=-0.183$), solar insolation--NDVI ($r=0.132$), and NDVI--NDWI ($r=-0.114$). Because the maximum absolute correlation was 0.627, no predictor pair exhibited a near-perfect linear relationship, and all eight variables were retained for subsequent modelling.

\section{Research Design and Experimental Protocol}
\label{sec:experimental-design}

\subsection{Overall Research Workflow}
\label{subsec:overall-workflow}

The experimental design was constructed to evaluate predictive discrimination, probability reliability, spatially informed partition robustness, and repeated-split stability under a common protocol. The workflow comprised the following stages: formation of modality-specific feature groups; transformation and training-only standardization of the predictors; spatially aware stratified partitioning; training of standalone baselines and EcoFuseNet-V2; generation of out-of-fold branch outputs for the hybrid framework; validation-based model selection and calibration; and final evaluation on an untouched test partition.

The training subset was used to estimate model parameters, preprocessing statistics, and class weights. The validation subset controlled checkpoint selection, meta-model selection, voting weights, temperature calibration, and other explicitly defined model-selection decisions. The test subset was withheld until the complete configuration had been fixed, thereby reducing selection bias in the final performance estimate \cite{cawley2010overfitting}. Calibrated EcoTreeFuseNet-Plus was treated as the principal proposed model, while the raw, bias-tuned, and hierarchy-aware variants were retained as secondary comparisons.

\subsection{Feature Groups}
\label{subsec:feature-groups}

Three feature configurations were defined to quantify the contribution of spectral and LiDAR-derived information under an identical experimental setting.

\subsubsection{Spectral-Only Group}
\label{subsubsec:spectral-only-group}

The spectral-only configuration contained NDVI and NDWI:

\begin{equation}
\mathbf{x}_{i}^{S}
=
\left[
x_{i}^{\mathrm{ndvi}},
x_{i}^{\mathrm{ndwi}}
\right]^{\mathsf{T}}
\in\mathbb{R}^{2}.
\label{eq:spectral-feature-group-design}
\end{equation}

This configuration measured the predictive information available from vegetation greenness and moisture-related spectral response without terrain or structural variables.

\subsubsection{LiDAR-Derived Group}
\label{subsubsec:lidar-only-group}

The LiDAR-derived configuration comprised elevation, slope, curvature, Topographic Position Index, solar insolation, and Canopy Height Model:

\begin{equation}
\mathbf{x}_{i}^{L}
=
\left[
x_{i}^{\mathrm{elev}},
x_{i}^{\mathrm{slope}},
x_{i}^{\mathrm{curv}},
x_{i}^{\mathrm{tpi}},
x_{i}^{\mathrm{solar}},
x_{i}^{\mathrm{chm}}
\right]^{\mathsf{T}}
\in\mathbb{R}^{6}.
\label{eq:lidar-feature-group-design}
\end{equation}

This group represented terrain configuration, environmental exposure, and vegetation structural height without spectral indices.

\subsubsection{Fused Group}
\label{subsubsec:fused-feature-group}

The fused configuration combined all eight predictors:

\begin{equation}
\mathbf{x}_{i}^{F}
=
\left[
\left(\mathbf{x}_{i}^{L}\right)^{\mathsf{T}},
\left(\mathbf{x}_{i}^{S}\right)^{\mathsf{T}}
\right]^{\mathsf{T}}
\in\mathbb{R}^{8}.
\label{eq:fused-feature-group-design}
\end{equation}

The fused group was used by the principal baseline comparisons and the proposed tree--neural framework. Within EcoFuseNet-V2, it was further separated into a five-variable terrain branch, a one-variable canopy branch, and a two-variable spectral branch, as described in Section~\ref{subsec:ecofusenet-v2}.

\subsection{Data Transformation and Standardization}
\label{subsec:data-transformation}

\subsubsection{Canopy-Height Transformation}
\label{subsubsec:chm-log-transformation}

The Canopy Height Model exhibited a sparse and strongly right-skewed distribution. It was therefore transformed before standardization using

\begin{equation}
x_{i,\mathrm{chm}}^{\log}
=
\log\left(1+x_{i,\mathrm{chm}}\right).
\label{eq:chm-log1p}
\end{equation}

The \texttt{log1p} transformation preserved valid zero values while compressing the influence of the relatively small number of large canopy-height observations. No equivalent nonlinear transformation was applied to the remaining predictors.

\subsubsection{Training-Only Standardization}
\label{subsubsec:feature-standardization}

For each feature $j$, the mean and standard deviation were estimated exclusively from the training partition:

\begin{align}
\mu_j
&=
\frac{1}{N_{\mathrm{tr}}}
\sum_{i\in\mathcal{D}_{\mathrm{tr}}}x_{ij},
\label{eq:training-feature-mean}
\\
\sigma_j
&=
\sqrt{
\frac{1}{N_{\mathrm{tr}}}
\sum_{i\in\mathcal{D}_{\mathrm{tr}}}
\left(x_{ij}-\mu_j\right)^2
}
+\varepsilon,
\label{eq:training-feature-sd}
\\
\widehat{x}_{ij}
&=
\frac{x_{ij}-\mu_j}{\sigma_j}.
\label{eq:training-standardization}
\end{align}

The same training-derived $\mu_j$ and $\sigma_j$ were then applied to the validation and test partitions. Validation and test observations therefore contributed no information to the preprocessing statistics. A corresponding training-only standardization procedure was later applied to the hybrid meta-feature matrix.

\subsection{Spatially Aware Stratified Partitioning}
\label{subsec:spatial-partitioning}

Ecological observations may exhibit spatial dependence, such that
nearby samples share environmental conditions and are consequently
more similar than geographically distant observations. Unconstrained
random splitting can therefore place closely related observations in
both training and test subsets and may yield optimistic estimates of
geographical generalization
\citep{roberts2017cross,valavi2019blockcv}. Conversely, imposing
strictly separated spatial blocks can substantially reduce class
coverage in a limited-sample problem containing 29 fine classes,
particularly for low-frequency categories.

A class-preserving, spatially informed partitioning strategy was
therefore adopted. Let
\(\mathbf{s}_i=(E_i,N_i)\) denote the projected UTM Easting and
Northing coordinates of observation \(i\). The projected study domain
was divided into a \(5\times5\) quantile grid by partitioning the
Easting and Northing coordinate distributions independently into five
quantile intervals. This produced 25 spatial blocks for the present
dataset. Each observation was assigned to the block defined by its
corresponding Easting- and Northing-bin indices. Spatial coordinates
were used only for partition construction and subsequent proximity
diagnostics; they were not included in the predictor matrix.

Within each spatial block, observations were allocated to training,
validation, and test subsets using fine-class stratification whenever
the available class counts permitted the requested subdivision. The
test subset was first separated using a nominal test fraction of
0.20. The remaining observations were then divided so that the
validation subset represented approximately 0.10 of the complete
dataset, corresponding to a validation fraction of
\(0.10/(0.70+0.10)=0.125\) within the remaining training--validation
pool.

When a spatial block did not contain sufficient observations or class
support for stratified subdivision, the implementation used a
seed-controlled random fallback within that block. Blocks containing
fewer than three observations were assigned entirely to the training
partition. For the primary seed-42 partition, one of the 25 spatial
blocks contained fewer than three observations. Among the remaining
24 blocks, stratified subdivision was feasible in 18 blocks, whereas
the random fallback was required in six blocks. No observations were
duplicated, interpolated, or synthetically generated during this
procedure.

The intended global allocation proportions were approximately

\begin{equation}
\left|\mathcal{D}_{\mathrm{tr}}\right|
\approx 0.70N,\qquad
\left|\mathcal{D}_{\mathrm{val}}\right|
\approx 0.10N,\qquad
\left|\mathcal{D}_{\mathrm{te}}\right|
\approx 0.20N,
\label{eq:outer-split-ratios}
\end{equation}

where \(N\) denotes the total number of observations. Because
allocation was performed independently within spatial blocks and was
constrained by the available class support, the realized proportions
could deviate slightly from these nominal targets through block-level
rounding.

After block-wise allocation, the training partition was checked to
ensure representation of every fine class. If a class had been absent
from the training subset, one observation of that class would have
been transferred from the validation subset, or from the test subset
if necessary, into the training subset. For the primary seed-42
partition, all 29 fine classes were already represented in the
training subset, so no such transfer was required.

The resulting subsets were mutually exclusive and collectively
exhaustive:

\begin{equation}
\mathcal{D}
=
\mathcal{D}_{\mathrm{tr}}
\cup
\mathcal{D}_{\mathrm{val}}
\cup
\mathcal{D}_{\mathrm{te}},
\qquad
\mathcal{D}_{a}\cap\mathcal{D}_{b}
=
\varnothing
\quad
(a\neq b).
\label{eq:disjoint-outer-splits}
\end{equation}

Seed 42 defined the primary partition used for the principal model
comparison. For the repeated-split analysis, the same \(5\times5\)
quantile-grid construction and allocation procedure was rerun
independently using seeds 42, 101, 202, 303, and 404. The spatial
partitioning logic and nominal allocation fractions were therefore
held constant, while seed-controlled within-block assignments were
allowed to vary. The resulting experiments assess sensitivity to both
data allocation and stochastic model training rather than repeatedly
evaluating an identical test subset.

This procedure should be interpreted as a spatially informed,
class-preserving partition rather than a strictly geographically
isolated or spatial-extrapolation design. Spatial separation was
therefore evaluated independently using the test-to-training
nearest-neighbour diagnostic reported in
Section~\ref{subsec:spatial-leakage-diagnostic}.

\subsection{Training, Validation, and Test Partitions}
\label{subsec:outer-partition-counts}

Seed 42 defined the primary outer partition used for the principal
model comparison. Following the spatially informed, class-preserving
partitioning procedure described in
Section~\ref{subsec:spatial-partitioning}, the 1,833 observations were
allocated to training, validation, and test subsets containing 1,273,
190, and 370 observations, respectively. The realized partition sizes
and proportions are reported in Table~\ref{tab:outer-split-counts}.

\begin{table}[!htb]
\centering
\caption{Training, validation, and test partition sizes for the
primary seed-42 experiment.}
\label{tab:outer-split-counts}
\begin{tabular}{lrr}
\toprule
\textbf{Partition} &
\textbf{Observations} &
\textbf{Percentage (\%)} \\
\midrule
Training   & 1,273 & 69.45 \\
Validation &   190 & 10.37 \\
Test       &   370 & 20.19 \\
\midrule
Total      & 1,833 & 100.00 \\
\bottomrule
\end{tabular}
\end{table}

The training subset was used to estimate model parameters,
training-derived preprocessing statistics, class weights, and
out-of-fold branch models. The validation subset was used exclusively
for model-selection decisions, including neural checkpoint selection,
meta-classifier selection, soft-voting weights, calibration
temperature, and the parameters of the secondary bias-tuned and
hierarchy-aware variants. The test subset remained untouched during
all preprocessing, training, model-selection, and calibration
decisions and was used only after the complete predictive
configuration had been fixed.

\begin{figure}[!ht]
    \centering
    \includegraphics[
        width=0.95\linewidth,
        height=0.55\textheight,
        keepaspectratio
    ]{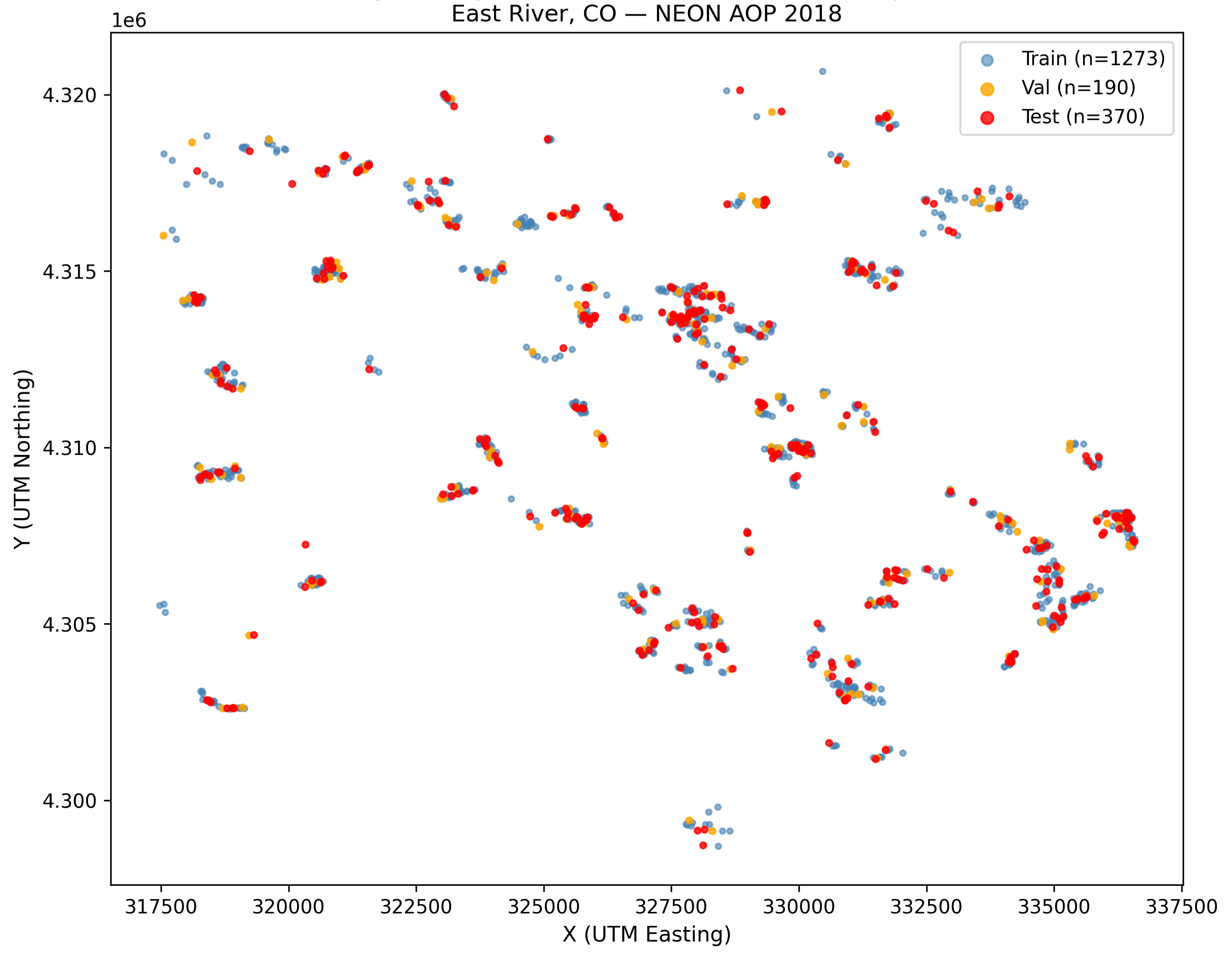}
    \caption{Spatial distribution of the training, validation, and
    test observations for the primary seed-42 partition in the East
    River study area. The partition contained 1,273 training,
    190 validation, and 370 test observations. The three subsets
    retained broad coverage of the study domain under the spatially
    informed, class-preserving partitioning strategy. The
    visualization documents the partition geometry and does not imply
    strict geographical isolation between training, validation, and
    test observations.}
    \label{fig:spatial-split}
\end{figure}

The spatial distribution of the three subsets is shown in
Figure~\ref{fig:spatial-split}. Training, validation, and test
observations were distributed across the study domain rather than
being concentrated in isolated portions of the coordinate space.
This pattern reflects the intended spatially informed,
class-preserving design and maintains geographical coverage for the
29-class classification problem. The figure is provided to document
the geometry of the primary partition and should not be interpreted
as evidence of strict geographical independence between the three
subsets.

Because spatially distributed subsets can still contain geographically
proximate observations, partition quality was additionally evaluated
using the test-to-training nearest-neighbour distance diagnostic
described in Section~\ref{subsec:spatial-leakage-diagnostic}.

\FloatBarrier

\subsection{Spatial Leakage Diagnostic}
\label{subsec:spatial-leakage-diagnostic}

Partition quality was examined using the Euclidean distance from each test observation to its nearest training observation in projected coordinate space:

\begin{equation}
d_{i}^{\mathrm{NN}}
=
\min_{j\in\mathcal{D}_{\mathrm{tr}}}
\left\|
\mathbf{s}_{i}^{\mathrm{te}}
-
\mathbf{s}_{j}^{\mathrm{tr}}
\right\|_{2},
\label{eq:spatial-nearest-neighbour}
\end{equation}

where $\mathbf{s}$ denotes the projected spatial coordinate. The implemented diagnostic raised a proximity flag when the median nearest-neighbour distance was below 1 m:

\begin{equation}
I_{\mathrm{prox}}
=
\mathbb{I}
\left[
\operatorname{median}
\left(d_{i}^{\mathrm{NN}}\right)
<1~\mathrm{m}
\right].
\label{eq:spatial-leakage-flag}
\end{equation}

For the seed-42 split, the median test-to-training nearest-neighbour distance was 36.86 m and the proximity flag was false. This result indicates that the partition did not contain widespread near-duplicate train--test locations under the implemented rule. It does not, however, establish complete spatial independence, and the split is therefore reported conservatively as spatially aware rather than strictly spatially isolated.

\subsection{Class-Weight Calculation}
\label{subsec:class-weighting}

Class weights were derived from the training labels only, preventing validation or test frequencies from influencing model fitting. Let $n_k$ denote the number of training observations in class $k$, and let

\begin{equation}
n_{\max}
=
\max_{1\leq k\leq K}n_k.
\end{equation}

The initial square-root inverse-frequency weight was

\begin{equation}
r_k
=
\sqrt{
\frac{n_{\max}}
{\max(n_k,1)}
}.
\label{eq:raw-class-weight}
\end{equation}

The weights were then normalized to have a mean of one:

\begin{equation}
w_k
=
\frac{r_k}
{\displaystyle
\frac{1}{K}
\sum_{j=1}^{K}r_j
}.
\label{eq:normalized-class-weight}
\end{equation}

The square-root form reduced majority-class dominance without applying the more aggressive amplification produced by full inverse-frequency weighting, a relevant consideration for small and imbalanced multiclass data \cite{he2009learning}. These weights were used in the class-balanced neural classification and prototype losses. Model-specific balancing options were used for conventional classifiers where supported.

\subsection{Reproducibility Controls}
\label{subsec:reproducibility-controls}

The complete experimental pipeline was implemented in Python, with
PyTorch used for EcoFuseNet-V2 and established tabular-learning
libraries used for the tree, boosting, and meta-classification
components. Numerical processing, statistical analysis, data
management, and visualization were performed using standard
scientific-computing libraries. Framework-level architecture,
optimization, out-of-fold, calibration, and secondary-variant
constants are summarized in
Table~\ref{tab:framework-hyperparameters}. Estimator-specific
hyperparameters, neural-training settings, and software versions were
fixed within the executable configuration and retained with the
reproducibility package.

Seed 42 was used for the primary experiment. Random states were
propagated, wherever supported, to the outer data partition,
conventional learners, neural initialization, mini-batch ordering,
out-of-fold partitioning, and validation-based ensemble search. The
complete calibrated pipeline was subsequently rerun using seeds
42, 101, 202, 303, and 404 to evaluate repeated-split stability. All
reported experiments were executed in a CPU environment without GPU
acceleration.

Preprocessing parameters were estimated from training data only. The
outer predictor scaler was fitted using the outer training partition,
whereas the hybrid meta-feature scaler was fitted using the
out-of-fold training meta-features. Validation data were used only for
predefined model-selection decisions, including neural checkpoint
selection, meta-classifier selection, soft-voting weights,
temperature calibration, and the secondary-variant parameters. Test
observations were not used for preprocessing, model selection, fusion
selection, or calibration.

For each experiment, the implementation retained the partition
metadata, validation-selected neural checkpoints, out-of-fold
features, refitted branch outputs, EcoFuseNet-V2 representations,
meta-feature scaler, selected meta-classifier, soft-voting weights,
calibration parameters, final probability arrays, predicted labels,
and statistical outputs. This saved-output design allowed the
principal metrics, tables, figures, and paired statistical analyses to
be regenerated from fixed predictions without retraining the complete
pipeline.

\subsection{Baseline Comparison Protocol}
\label{subsec:baseline-comparison-protocol}

All standalone baselines, neural variants, modality ablations, and proposed-model variants used the same seed-specific outer training, validation, and test observations. Target encoding and predictor definitions were also held constant. Consequently, performance differences within a seed were not confounded by alternative test samples.

Model fitting and hyperparameter or checkpoint selection were restricted to the training and validation partitions. Test labels were not used to choose a baseline, fusion strategy, calibration temperature, class adjustment, or hierarchy parameter. The frozen models produced class probabilities and labels for the same test observations, enabling paired statistical comparisons.

The main statistical reference was Calibrated EcoTreeFuseNet-Plus. Raw EcoTreeFuseNet-Plus, bias-tuned EcoTreeFuseNet-Plus, and hierarchy-aware EcoTreeFuseNet-Plus were treated as secondary variants rather than alternative primary models.

\subsection{Evaluation Metrics}
\label{subsec:evaluation-metrics}

Let $K=29$ denote the number of fine classes and $N$ the number of
test observations. For class $k$, precision, recall, and F1-score were
defined as

\begin{align}
P_k
&=
\frac{TP_k}{TP_k+FP_k},
\\
R_k
&=
\frac{TP_k}{TP_k+FN_k},
\\
F1_k
&=
\frac{2P_kR_k}{P_k+R_k}.
\label{eq:classwise-f1}
\end{align}

Undefined divisions were assigned zero consistently with the
evaluation implementation.

\subsubsection{Discrimination Metrics}
\label{subsubsec:discrimination-metrics}

Overall accuracy measured the proportion of correctly classified
observations:

\begin{equation}
\mathrm{Accuracy}
=
\frac{1}{N}
\sum_{i=1}^{N}
\mathbb{I}\!\left(\widehat{y}_i=y_i\right).
\label{eq:accuracy}
\end{equation}

Macro F1 assigned equal importance to each of the $K$ classes:

\begin{equation}
\mathrm{MacroF1}
=
\frac{1}{K}
\sum_{k=1}^{K}F1_k,
\label{eq:macro-f1}
\end{equation}

whereas weighted F1 accounted for the test-set support $n_k$ of each
class:

\begin{equation}
\mathrm{WeightedF1}
=
\sum_{k=1}^{K}
\frac{n_k}{N}F1_k.
\label{eq:weighted-f1}
\end{equation}

Balanced accuracy was defined as the mean recall across classes:

\begin{equation}
\mathrm{BA}
=
\frac{1}{K}
\sum_{k=1}^{K}R_k.
\label{eq:balanced-accuracy}
\end{equation}

Macro F1 and balanced accuracy were emphasized because they reduce
the dominance of high-frequency classes in imbalanced multiclass
evaluation
\citep{sokolova2009systematic,brodersen2010balanced}.

For the multiclass confusion matrix $\mathbf{M}$, let $r_k$ and $c_k$
denote the true and predicted marginal totals for class $k$,
respectively, and let
$n=\sum_{ij}M_{ij}=N$. The multiclass Matthews correlation
coefficient was calculated as

\begin{equation}
\mathrm{MCC}
=
\frac{
n\sum_{k}M_{kk}
-
\sum_{k}r_kc_k
}{
\sqrt{
\left(n^2-\sum_{k}c_k^2\right)
\left(n^2-\sum_{k}r_k^2\right)
}
}.
\label{eq:multiclass-mcc}
\end{equation}

MCC summarizes the complete multiclass confusion structure and
remains informative when class frequencies are unequal
\citep{gorodkin2004comparing}.

Cohen's kappa was calculated as

\begin{equation}
\kappa
=
\frac{p_o-p_e}{1-p_e},
\qquad
p_o=\frac{\sum_k M_{kk}}{N},
\qquad
p_e=\frac{\sum_k r_kc_k}{N^2},
\label{eq:cohens-kappa}
\end{equation}

where $p_o$ denotes the observed agreement and $p_e$ denotes the
agreement expected from the true and predicted marginal
distributions \citep{cohen1960coefficient}.

\subsubsection{Probability-Reliability Metrics}
\label{subsubsec:probability-metrics}

For predicted class probabilities $p_{ik}$, the maximum predictive
confidence and correctness indicator for observation $i$ were defined
as

\begin{equation}
q_i=\max_k p_{ik},
\qquad
a_i=\mathbb{I}\!\left(\widehat{y}_i=y_i\right).
\label{eq:confidence-correctness}
\end{equation}

Expected calibration error (ECE) was computed using $B=10$
equal-width confidence bins over the interval $[0,1]$. Consistent with
the fixed implementation, the bins were left-closed and right-open:

\begin{equation}
I_b
=
\left\{
i:
\frac{b-1}{10}
\le q_i <
\frac{b}{10}
\right\},
\qquad
b=1,\ldots,10.
\label{eq:ece-bin-definition}
\end{equation}

Thus, the implemented intervals were
$[0,0.1),[0.1,0.2),\ldots,[0.9,1.0)$. A confidence value exactly equal
to $1.0$ was therefore not assigned to a bin under the implemented
convention. The same binning rule was applied consistently to every
model and calibration state.

For each non-empty bin, empirical accuracy and mean confidence were

\begin{equation}
\operatorname{acc}(I_b)
=
\frac{1}{|I_b|}
\sum_{i\in I_b}a_i,
\qquad
\operatorname{conf}(I_b)
=
\frac{1}{|I_b|}
\sum_{i\in I_b}q_i.
\label{eq:ece-bin-statistics}
\end{equation}

ECE was then calculated as

\begin{equation}
\mathrm{ECE}
=
\sum_{b=1}^{10}
\frac{|I_b|}{N}
\left|
\operatorname{acc}(I_b)
-
\operatorname{conf}(I_b)
\right|.
\label{eq:ece}
\end{equation}

Empty bins contributed zero to the summation. Lower ECE indicates
closer agreement between predictive confidence and observed
correctness \citep{guo2017calibration}.

The multiclass Brier score evaluated the complete predicted
probability vector:

\begin{equation}
\mathrm{Brier}
=
\frac{1}{N}
\sum_{i=1}^{N}
\sum_{k=1}^{K}
\left(
p_{ik}-\mathbb{I}[y_i=k]
\right)^2.
\label{eq:brier-score}
\end{equation}

The implementation used the summed multiclass form above and did not
divide the per-observation squared error by $K$.

Negative log-likelihood was computed from the probability assigned to
the true class:

\begin{equation}
\mathrm{NLL}
=
-\frac{1}{N}
\sum_{i=1}^{N}
\log\!\left(
\max\{p_{i,y_i},10^{-8}\}
\right).
\label{eq:nll}
\end{equation}

The lower clipping value of $10^{-8}$ prevented numerical
singularities when evaluating probabilities approaching zero. Both
the Brier score and NLL are proper probabilistic measures that assess
the complete predictive distribution, whereas ECE summarizes
confidence agreement within bins
\citep{brier1950verification,guo2017calibration,
gneiting2007strictly}. Lower values indicate better probability
quality for all three reliability metrics.

\subsection{Statistical Evaluation}
\label{subsec:statistical-evaluation}

Statistical analyses were designed to quantify uncertainty in the
principal performance estimates, compare models on identical test
observations, and evaluate the diagnostic behaviour of the
EcoFuseNet-V2 neural branch. Unless otherwise stated, the primary
statistical analyses used the seed-42 test partition.

\subsubsection{Bootstrap Confidence Intervals}
\label{subsubsec:bootstrap-confidence-intervals}

Uncertainty in the principal discrimination metrics was estimated
using $B=1000$ bootstrap resamples of the seed-42 test observations.
Each bootstrap sample contained $N_{\mathrm{te}}$ indices drawn with
replacement from the original test set. For metric $m$, the
$b$th bootstrap estimate was

\begin{equation}
m^{(b)}
=
m\!\left(
\mathbf{y}_{I_b},
\widehat{\mathbf{y}}_{I_b}
\right),
\qquad
b=1,\ldots,B,
\label{eq:bootstrap-metric}
\end{equation}

where $I_b$ denotes the resampled test indices. The bootstrap
procedure used a fixed random seed of 42 for reproducibility.

A two-sided 95\% percentile confidence interval was calculated as

\begin{equation}
\mathrm{CI}_{95\%}(m)
=
\left[
Q_{0.025}\!\left(m^{(1)},\ldots,m^{(B)}\right),
Q_{0.975}\!\left(m^{(1)},\ldots,m^{(B)}\right)
\right],
\label{eq:bootstrap-ci}
\end{equation}

where $Q_{\alpha}$ denotes the empirical $\alpha$-quantile
\citep{efron1979bootstrap}. Confidence intervals were reported for
macro F1 and balanced accuracy, the two principal class-balanced
discrimination measures.

\subsubsection{Paired Bootstrap Comparisons}
\label{subsubsec:paired-bootstrap-tests}

Because competing models generated predictions for the same test
observations, model comparisons were performed using paired bootstrap
resampling. For every bootstrap iteration, identical resampled indices
were applied to both models. The metric difference was therefore

\begin{equation}
\Delta^{(b)}
=
m_A^{(b)}-m_B^{(b)},
\qquad
b=1,\ldots,B,
\label{eq:paired-bootstrap-difference}
\end{equation}

where model $A$ was Calibrated EcoTreeFuseNet-Plus and model $B$ was
the corresponding baseline, ablation, or secondary variant. The
observed difference on the original test set was

\begin{equation}
\Delta_{\mathrm{obs}}
=
m_A-m_B.
\label{eq:paired-bootstrap-observed}
\end{equation}

The 95\% confidence interval for the paired difference was obtained
from the 2.5th and 97.5th percentiles of
$\{\Delta^{(b)}\}_{b=1}^{B}$. The implementation used
small-sample-corrected bootstrap tail probabilities. For the
directional alternative $A>B$, the one-sided probability was

\begin{equation}
p_{\mathrm{one}}
=
\frac{
1+\sum_{b=1}^{B}
\mathbb{I}\!\left(\Delta^{(b)}\leq0\right)
}{
B+1
},
\label{eq:paired-bootstrap-pone}
\end{equation}

and the reported two-sided probability was

\begin{equation}
p_{\mathrm{two}}
=
\min\!\left\{
1,\;
2\min\!\left[
\frac{
1+\sum_{b=1}^{B}
\mathbb{I}\!\left(\Delta^{(b)}\leq0\right)
}{
B+1
},
\frac{
1+\sum_{b=1}^{B}
\mathbb{I}\!\left(\Delta^{(b)}\geq0\right)
}{
B+1
}
\right]
\right\}.
\label{eq:paired-bootstrap-ptwo}
\end{equation}

Pairing preserves the dependence between predictions made on the same
observations and is therefore preferable to treating model-level
scores as independent. The reported model-comparison $p$-values were
not adjusted for multiplicity and are interpreted as model-specific
evidence rather than simultaneous family-wise tests.

\subsubsection{McNemar Test}
\label{subsubsec:mcnemar-test}

Pairwise differences in test-set correctness were additionally
examined using a two-sided McNemar test
\citep{mcnemar1947note}. Let model $A$ denote Calibrated
EcoTreeFuseNet-Plus and model $B$ the comparator. The discordant
counts were defined as

\begin{equation}
n_{01}
=
\sum_{i=1}^{N_{\mathrm{te}}}
\mathbb{I}
\left(
\widehat{y}_{A,i}=y_i,\;
\widehat{y}_{B,i}\neq y_i
\right),
\label{eq:mcnemar-n01}
\end{equation}

and

\begin{equation}
n_{10}
=
\sum_{i=1}^{N_{\mathrm{te}}}
\mathbb{I}
\left(
\widehat{y}_{A,i}\neq y_i,\;
\widehat{y}_{B,i}=y_i
\right).
\label{eq:mcnemar-n10}
\end{equation}

When the total number of discordant observations satisfied

\begin{equation}
n_{01}+n_{10}<25,
\end{equation}

the implementation used the exact two-sided binomial form of
McNemar's test. Otherwise, the continuity-corrected chi-square
statistic was

\begin{equation}
\chi^2
=
\frac{
\left(
|n_{01}-n_{10}|-1
\right)^2
}{
n_{01}+n_{10}
},
\label{eq:mcnemar-statistic}
\end{equation}

with one degree of freedom. If no discordant observations were
present, the comparison was assigned $p=1$. McNemar's test therefore
complemented metric-level bootstrap comparisons with an
observation-level assessment of paired classification correctness.

\subsubsection{Neural-Branch Diagnostic Analyses}
\label{subsubsec:neural-diagnostic-methods}

Additional analyses were performed to characterize the internal
behaviour and uncertainty signal of EcoFuseNet-V2. These diagnostics
were exploratory analyses of the standalone neural branch and were
not used to select or modify the final calibrated
EcoTreeFuseNet-Plus predictions.

\paragraph{Gate--CHM association.}

For each test observation, the 64-dimensional gate representation
$\mathbf{g}_i$ was reduced to a scalar by averaging across its gate
dimensions:

\begin{equation}
\bar{g}_i
=
\frac{1}{d_f}
\sum_{j=1}^{d_f}g_{ij},
\qquad
d_f=64.
\label{eq:scalar-gate}
\end{equation}

Ecological validation was performed using the true fine-class labels.
For each class $k$ represented in the test set, the class-level mean
gate value was

\begin{equation}
\bar{g}_k
=
\frac{1}{n_k}
\sum_{i:y_i=k}
\bar{g}_i,
\label{eq:class-mean-gate}
\end{equation}

the mean raw canopy height was

\begin{equation}
\overline{\mathrm{CHM}}_k
=
\frac{1}{n_k}
\sum_{i:y_i=k}
\mathrm{CHM}_i,
\label{eq:class-mean-chm}
\end{equation}

and the proportion of observations with positive CHM was

\begin{equation}
r^{+}_{k}
=
\frac{1}{n_k}
\sum_{i:y_i=k}
\mathbb{I}\!\left(\mathrm{CHM}_i>0\right).
\label{eq:class-positive-chm}
\end{equation}

Spearman rank correlations were then calculated across represented
classes between
$\bar{g}_k$ and $\overline{\mathrm{CHM}}_k$, and between
$\bar{g}_k$ and $r^{+}_{k}$. The predefined ecological
gate-validity criterion required a positive and statistically
significant association between class-level mean gate value and mean
CHM at the conventional $p<0.05$ level. The positive-CHM-rate
correlation was retained as an additional diagnostic because the CHM
variable was highly zero-inflated.

\paragraph{Prototype compactness diagnostic.}

Prototype quality was assessed using the deterministic
EcoFuseNet-V2 test embeddings and the learned fine-class prototypes.
For each class $k$, the within-class radius was defined as the mean
Euclidean distance between test embeddings belonging to class $k$ and
their corresponding learned prototype:

\begin{equation}
R^{\mathrm{intra}}_k
=
\frac{1}{n_k}
\sum_{i:y_i=k}
\left\|
\mathbf{e}_i-\mathbf{P}_k
\right\|_2.
\label{eq:prototype-intra}
\end{equation}

The nearest inter-prototype distance for class $k$ was

\begin{equation}
R^{\mathrm{inter}}_k
=
\min_{j\neq k}
\left\|
\mathbf{P}_k-\mathbf{P}_j
\right\|_2.
\label{eq:prototype-inter}
\end{equation}

A class-specific compactness ratio was then calculated as

\begin{equation}
\mathrm{CR}_k
=
\frac{
R^{\mathrm{inter}}_k
}{
R^{\mathrm{intra}}_k+10^{-8}
},
\label{eq:prototype-cr}
\end{equation}

and the reported macro compactness ratio was

\begin{equation}
\mathrm{CR}_{\mathrm{macro}}
=
\frac{1}{K}
\sum_{k=1}^{K}
\mathrm{CR}_k.
\label{eq:prototype-cr-macro}
\end{equation}

The reported mean within-class distance was the arithmetic mean of
$R^{\mathrm{intra}}_k$ across the 29 classes, whereas the reported
between-prototype distance was the mean of the nearest
inter-prototype distances $R^{\mathrm{inter}}_k$. These quantities
were treated as exploratory representation diagnostics rather than
as measures of final-model feature contribution.

\paragraph{MC-dropout predictive uncertainty.}

Predictive uncertainty for EcoFuseNet-V2 was estimated using
Monte Carlo dropout. For the standard experiment, dropout was kept
active during inference and $T_{\mathrm{MC}}=50$ stochastic forward
passes were performed for every seed-42 test observation. Let
$p^{(t)}_{ik}$ denote the predicted probability for class $k$ in
stochastic pass $t$. The MC-averaged probability was

\begin{equation}
\bar{p}_{ik}
=
\frac{1}{T_{\mathrm{MC}}}
\sum_{t=1}^{T_{\mathrm{MC}}}
p^{(t)}_{ik},
\qquad
T_{\mathrm{MC}}=50.
\label{eq:mc-mean-probability}
\end{equation}

Predictive entropy was calculated in natural-logarithm units as

\begin{equation}
H_i
=
-
\sum_{k=1}^{K}
\bar{p}_{ik}
\log\!\left(
\bar{p}_{ik}+10^{-8}
\right).
\label{eq:mc-predictive-entropy}
\end{equation}

Test observations were then separated according to whether the
standalone EcoFuseNet-V2 prediction was correct or incorrect. The
prespecified directional hypothesis was

\begin{equation}
H_1:
H_{\mathrm{incorrect}}
>
H_{\mathrm{correct}},
\label{eq:entropy-directional-hypothesis}
\end{equation}

reflecting the expectation that incorrect neural predictions would,
on average, exhibit greater predictive uncertainty.

The two entropy distributions were compared using a one-sided
Mann--Whitney $U$ test with the alternative that the entropy of
incorrect predictions was greater than that of correct predictions.
A standardized mean difference was additionally calculated using the
same implementation as the reported results. Let
$\bar{H}_{I}$ and $\bar{H}_{C}$ denote the mean entropies of
incorrect and correct predictions, respectively, and let
$s_I$ and $s_C$ denote their population standard deviations
(i.e., calculated with the $n$ denominator). Cohen's $d$ was defined
as

\begin{equation}
d
=
\frac{
\bar{H}_{I}-\bar{H}_{C}
}{
\sqrt{
\left(
s_I^2+s_C^2
\right)/2
}
+10^{-8}
}.
\label{eq:entropy-cohens-d}
\end{equation}

This definition uses the square root of the average of the two
population variances rather than the conventional pooled
sample-variance estimator. It is stated explicitly so that the
reported effect size is reproducible from the saved entropy arrays.
The MC-dropout uncertainty analysis applies exclusively to
EcoFuseNet-V2 and should not be interpreted as an uncertainty
estimate for Calibrated EcoTreeFuseNet-Plus.

\subsection{Repeated-Seed and Split Stability}
\label{subsec:repeated-seed-protocol}

The complete proposed-model pipeline was repeated independently using the seeds

\begin{equation}
\mathcal{S}
=
\{42,101,202,303,404\}.
\label{eq:repeated-seeds}
\end{equation}

For each seed, the spatially aware partition, model initialization, out-of-fold generation, validation-based source selection, calibration, and test evaluation were rerun. This protocol assessed sensitivity to both data partitioning and stochastic training rather than repeating only the final classifier on a fixed split.

For metric $R_s$ obtained under seed $s$, stability was summarized using

\begin{align}
\overline{R}
&=
\frac{1}{S}
\sum_{s=1}^{S}R_s,
\\
s_R
&=
\sqrt{
\frac{1}{S-1}
\sum_{s=1}^{S}
\left(R_s-\overline{R}\right)^2
},
\qquad S=5.
\label{eq:repeated-seed-summary}
\end{align}

Results were reported as mean $\pm$ sample standard deviation for the principal discrimination and calibration metrics. Because the outer test membership and ordering could differ across seeds, these runs were interpreted as repeated-seed/split stability evidence rather than as a probability ensemble.

\section{Proposed Calibrated EcoTreeFuseNet-Plus Framework}
\label{sec:proposed-framework}

\subsection{Framework Overview}
\label{subsec:framework-overview}

\begin{figure}[!ht]
    \centering
    \includegraphics[
        width=0.98\linewidth,
        height=0.52\textheight,
        keepaspectratio
    ]{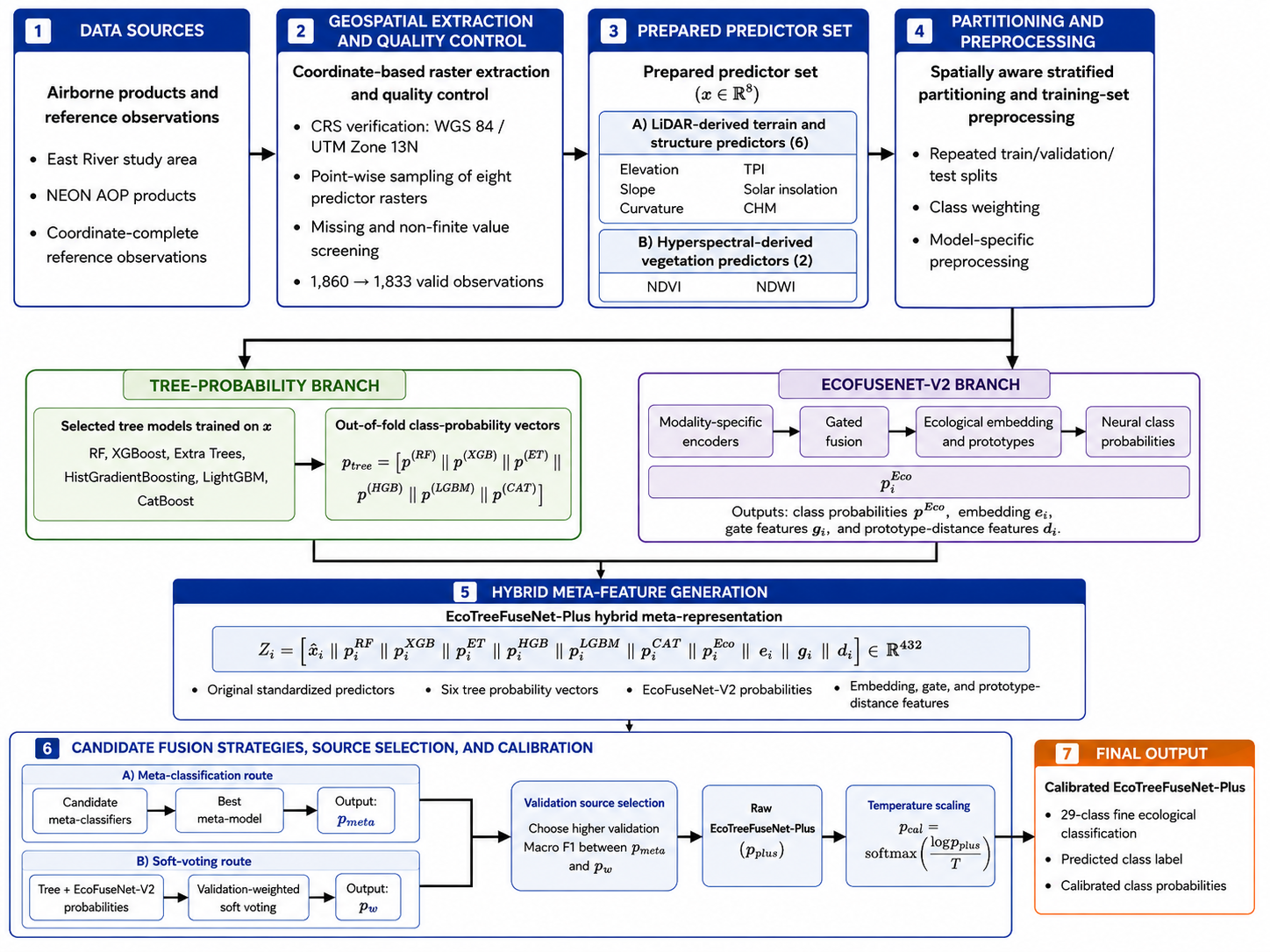}
    \caption{Overview of the proposed Calibrated
    EcoTreeFuseNet-Plus framework. Coordinate-based raster extraction
    and quality control produce eight predictors from LiDAR-derived
    terrain and canopy products and hyperspectral-derived vegetation
    indices. Following spatially informed partitioning and
    training-set preprocessing, parallel tree-probability and
    EcoFuseNet-V2 branches generate the inputs to the
    432-dimensional hybrid meta-representation. Candidate
    meta-classification and validation-weighted soft-voting routes are
    evaluated in parallel, and the source with the higher validation
    macro F1 defines the raw EcoTreeFuseNet-Plus probabilities. The
    selected probability source is subsequently temperature-scaled to
    generate calibrated probabilities and final predictions for the
    29 fine ecological classes.}
    \label{fig:proposed-framework}
\end{figure}

Calibrated EcoTreeFuseNet-Plus was developed as a hybrid
probability-fusion framework for fine-grained ecological classification
from small-sample tabular data. The framework integrates
coordinate-based geospatial feature extraction, complementary
LiDAR-derived and hyperspectral-derived predictors, tree-ensemble
probability branches, the EcoFuseNet-V2 neural branch, leakage-aware
out-of-fold feature generation, validation-based fusion selection, and
post-hoc temperature scaling. The complete methodological framework is
illustrated in Figure~\ref{fig:proposed-framework}.

For observation $i$, the standardized predictor vector was partitioned
as

\begin{equation}
\widehat{\mathbf{x}}_{i}
=
\left[
\left(\widehat{\mathbf{x}}_{i}^{T}\right)^{\mathsf{T}},
\left(\widehat{\mathbf{x}}_{i}^{C}\right)^{\mathsf{T}},
\left(\widehat{\mathbf{x}}_{i}^{S}\right)^{\mathsf{T}}
\right]^{\mathsf{T}},
\label{eq:framework-input-partition}
\end{equation}

where

\begin{align}
\widehat{\mathbf{x}}_{i}^{T}
&=
[
\mathrm{elevation},
\mathrm{slope},
\mathrm{curvature},
\mathrm{TPI},
\mathrm{solar\ insolation}
]^{\mathsf{T}}
\in\mathbb{R}^{5},
\\
\widehat{\mathbf{x}}_{i}^{C}
&=
[
\log(1+\mathrm{CHM})
]
\in\mathbb{R}^{1},
\\
\widehat{\mathbf{x}}_{i}^{S}
&=
[
\mathrm{NDVI},
\mathrm{NDWI}
]^{\mathsf{T}}
\in\mathbb{R}^{2}.
\end{align}

The principal output of the complete framework is the calibrated
fine-class probability distribution. Raw EcoTreeFuseNet-Plus,
Bias-Tuned EcoTreeFuseNet-Plus, and Hierarchy-Aware
EcoTreeFuseNet-Plus were retained as secondary variants for comparison
and sensitivity analysis.

\FloatBarrier

\subsection{EcoFuseNet-V2 Neural Branch}
\label{subsec:ecofusenet-v2}

\subsubsection{Modality-Specific Encoders and Gated Fusion}
\label{subsubsec:modality-gated-fusion}

EcoFuseNet-V2 processes terrain, canopy, and spectral predictors through separate encoders. The terrain and spectral pathways use compact multilayer representations comprising linear projection, layer normalization, GELU activation, and dropout. Because CHM was highly sparse, its structural encoder was deliberately kept low-capacity and did not use dropout.

The three encoded representations were

\begin{equation}
\mathbf{h}_{i}^{T}
=
f_{T}\!\left(\widehat{\mathbf{x}}_{i}^{T}\right),
\qquad
\mathbf{h}_{i}^{C}
=
f_{C}\!\left(\widehat{\mathbf{x}}_{i}^{C}\right),
\qquad
\mathbf{h}_{i}^{S}
=
f_{S}\!\left(\widehat{\mathbf{x}}_{i}^{S}\right),
\label{eq:modality-encoders}
\end{equation}

where

\begin{equation}
\mathbf{h}_{i}^{T},
\mathbf{h}_{i}^{C},
\mathbf{h}_{i}^{S}
\in\mathbb{R}^{d_f},
\qquad
d_f=64.
\end{equation}

A dimension-wise gate was computed from the concatenated modality representations:

\begin{equation}
\mathbf{g}_{i}
=
\sigma
\left(
\operatorname{GateNet}
\left[
\mathbf{h}_{i}^{C};
\mathbf{h}_{i}^{T};
\mathbf{h}_{i}^{S}
\right]
\right)
\in(0,1)^{d_f},
\label{eq:modality-gate}
\end{equation}

where $\sigma(\cdot)$ denotes the sigmoid function. The initial fused representation was then obtained as

\begin{equation}
\mathbf{z}_{0,i}
=
\mathbf{g}_{i}\odot\mathbf{h}_{i}^{T}
+
\left(\mathbf{1}-\mathbf{g}_{i}\right)
\odot\mathbf{h}_{i}^{S}
+
\alpha_{\mathrm{CHM}}\mathbf{h}_{i}^{C},
\label{eq:ecofusenet-gated-fusion}
\end{equation}

where $\odot$ denotes element-wise multiplication. The learnable scalar $\alpha_{\mathrm{CHM}}$ was initialized at 0.05 so that the sparse CHM representation could contribute without dominating the terrain--spectral fusion.

The fused vector was refined through a residual transformation:

\begin{equation}
\mathbf{z}_{i}
=
\operatorname{LN}
\left[
\mathbf{z}_{0,i}
+
\operatorname{FusionRefine}
\left(
\operatorname{LN}(\mathbf{z}_{0,i})
\right)
\right],
\label{eq:residual-fusion-refinement}
\end{equation}

where $\operatorname{LN}(\cdot)$ denotes layer normalization.

\subsubsection{Ecological Embedding and Fine-Class Prediction}
\label{subsubsec:embedding-classification}

The refined representation was projected into a 128-dimensional ecological embedding:

\begin{equation}
\mathbf{e}_{i}
=
\operatorname{LN}
\left(
\mathbf{W}_{e}\mathbf{z}_{i}
+
\mathbf{b}_{e}
\right)
\in\mathbb{R}^{d_e},
\qquad
d_e=128.
\label{eq:ecological-embedding}
\end{equation}

Fine-class logits and probabilities were calculated as

\begin{align}
\mathbf{a}_{i}
&=
\mathbf{W}_{y}\mathbf{e}_{i}
+
\mathbf{b}_{y}
\in\mathbb{R}^{K},
\\
p_{ik}^{\mathrm{Eco}}
&=
\frac{\exp(a_{ik})}
{\sum_{r=1}^{K}\exp(a_{ir})},
\qquad
K=29.
\label{eq:ecofusenet-probability}
\end{align}

The standalone EcoFuseNet-V2 prediction was

\begin{equation}
\widehat{y}_{i}^{\mathrm{Eco}}
=
\arg\max_{k}p_{ik}^{\mathrm{Eco}}.
\end{equation}

\subsubsection{Fine and Coarse Prototype Representation}
\label{subsubsec:prototype-representation}

EcoFuseNet-V2 additionally maintained a fine-class prototype
$\mathbf{P}_{k}\in\mathbb{R}^{d_e}$ for each of the $K=29$ classes and a coarse prototype
$\mathbf{Q}_{c}\in\mathbb{R}^{d_e}$ for each of the $C=5$ ecological groups. Prototype-based representations provide an auxiliary geometric structure in which observations can be compared with class centres \cite{snell2017prototypical}.

The squared distance between observation $i$ and fine prototype $k$ was

\begin{equation}
d_{ik}
=
\left\|
\mathbf{e}_{i}-\mathbf{P}_{k}
\right\|_{2}^{2}.
\label{eq:prototype-distance}
\end{equation}

A strictly positive prototype temperature was defined as

\begin{equation}
\tau
=
\tau_{\min}
+
\operatorname{softplus}
\left(
\tau_{\mathrm{raw}}
\right),
\qquad
\tau_{\min}=0.1,
\label{eq:prototype-temperature}
\end{equation}

giving distance-based logits

\begin{equation}
s_{ik}
=
-\frac{d_{ik}}{\tau}.
\label{eq:prototype-logits}
\end{equation}

After the initial cross-entropy warm-up, the fine prototypes were initialized from training embeddings:

\begin{equation}
\mathbf{P}_{k}
\leftarrow
\frac{1}{n_k}
\sum_{\substack{i\in\mathcal{D}_{\mathrm{tr}}\\y_i=k}}
\mathbf{e}_{i},
\label{eq:fine-prototype-initialization}
\end{equation}

and the coarse prototypes were initialized from the corresponding fine prototypes:

\begin{equation}
\mathbf{Q}_{c}
\leftarrow
\frac{1}{|\mathcal{K}_{c}|}
\sum_{k\in\mathcal{K}_{c}}
\mathbf{P}_{k},
\qquad
\mathcal{K}_{c}
=
\{k:\phi(k)=c\},
\label{eq:coarse-prototype-initialization}
\end{equation}

where $\phi(k)$ maps fine class $k$ to its coarse ecological category. All available training embeddings were used for classes with limited representation.

EcoFuseNet-V2 exported four quantities for subsequent fusion:

\begin{equation}
\left\{
\mathbf{p}_{i}^{\mathrm{Eco}},
\mathbf{e}_{i},
\mathbf{g}_{i},
\mathbf{d}_{i}
\right\},
\label{eq:ecofusenet-exported-features}
\end{equation}

corresponding to 29 class probabilities, a 128-dimensional embedding, a 64-dimensional gate vector, and 29 prototype distances.

\subsection{EcoFuseNet-V2 Objective and Training}
\label{subsec:ecofusenet-training}

\subsubsection{Learning Objective}
\label{subsubsec:ecofusenet-loss}

Class imbalance was handled using the training-derived weights defined in Section~\ref{subsec:class-weighting}. With label-smoothing parameter $\varepsilon$, the smoothed target was

\begin{equation}
q_{ik}
=
(1-\varepsilon)
\mathbb{I}[k=y_i]
+
\frac{\varepsilon}{K},
\qquad
\varepsilon=0.03.
\label{eq:label-smoothed-target}
\end{equation}

Label smoothing was used to reduce excessively concentrated target distributions during neural training \cite{szegedy2016rethinking}. The weighted label-smoothed cross-entropy was

\begin{equation}
\mathcal{L}_{\mathrm{CE}}
=
-\frac{1}{B}
\sum_{i=1}^{B}
w_{y_i}
\sum_{k=1}^{K}
q_{ik}
\log p_{ik}^{\mathrm{Eco}},
\label{eq:weighted-label-smoothed-ce}
\end{equation}

where $B$ denotes the mini-batch size.

The prototype contrastive loss was

\begin{equation}
\mathcal{L}_{\mathrm{proto}}
=
\frac{1}{B}
\sum_{i=1}^{B}
w_{y_i}
\left[
-s_{i,y_i}
+
\log
\left(
\sum_{k=1}^{K}\exp(s_{ik})
\right)
\right].
\label{eq:prototype-contrastive-loss}
\end{equation}

The coarse-consistency loss encouraged embeddings to remain close to the prototype of their correct coarse category:

\begin{equation}
\mathcal{L}_{\mathrm{coarse}}
=
\frac{1}{B}
\sum_{i=1}^{B}
\left\|
\mathbf{e}_{i}
-
\mathbf{Q}_{c_i}
\right\|_{2}^{2}.
\label{eq:coarse-consistency-loss}
\end{equation}

A bounded margin loss was used to discourage collapsed fine prototypes:

\begin{equation}
\mathcal{L}_{\mathrm{sep}}
=
\frac{2}{K(K-1)}
\sum_{1\leq j<k\leq K}
\max
\left(
0,
m-
\left\|
\mathbf{P}_{j}-\mathbf{P}_{k}
\right\|_{2}
\right)^{2},
\label{eq:prototype-separation-loss}
\end{equation}

where $m=1.0$. The complete EcoFuseNet-V2 objective was

\begin{equation}
\mathcal{L}_{\mathrm{V2}}
=
\mathcal{L}_{\mathrm{CE}}
+
\lambda_{p}\mathcal{L}_{\mathrm{proto}}
+
\lambda_{c}\mathcal{L}_{\mathrm{coarse}}
+
\lambda_{s}\mathcal{L}_{\mathrm{sep}}.
\label{eq:complete-ecofusenet-loss}
\end{equation}

Cross-entropy remained the dominant objective, while the prototype terms acted as weak auxiliary regularizers.

\subsubsection{Two-Stage Optimization}
\label{subsubsec:two-stage-training}

Training proceeded in two stages. First, EcoFuseNet-V2 was optimized using only $\mathcal{L}_{\mathrm{CE}}$ so that the embedding and fine-class decision boundaries could stabilize before prototype constraints were introduced. Fine and coarse prototypes were then initialized from the resulting training embeddings. In the second stage, optimization continued using the complete objective in Eq.~\eqref{eq:complete-ecofusenet-loss}.

AdamW was used for neural optimization \cite{loshchilov2019decoupled}. The gradient norm was clipped at 1.0:

\begin{equation}
\left\|
\nabla_{\boldsymbol{\theta}}
\mathcal{L}_{\mathrm{V2}}
\right\|_{2}
\leq 1.0.
\label{eq:gradient-clipping}
\end{equation}

Validation macro F1 controlled learning-rate scheduling, checkpoint selection, and early stopping. After training, the checkpoint with the highest validation macro F1 was restored before feature extraction or test inference. The test partition was not used for neural checkpoint selection.

For the full EcoFuseNet-V2 model,

\begin{equation}
(\lambda_p,\lambda_c,\lambda_s)
=
(0.03,0.01,0.001),
\label{eq:full-neural-loss-weights}
\end{equation}

whereas the fold-specific neural models used for out-of-fold feature generation employed weaker regularization:

\begin{equation}
(\lambda_p,\lambda_c,\lambda_s)_{\mathrm{OOF}}
=
(0.01,0.005,0.0005).
\label{eq:oof-neural-loss-weights}
\end{equation}

\subsection{Tree Probability Branches and Out-of-Fold Construction}
\label{subsec:tree-oof-construction}

\subsubsection{Tree-Ensemble Branches}
\label{subsubsec:tree-probability-branches}

Six tree-based learners were used as probability-generating branches: Random Forest, XGBoost, ExtraTrees, HistGradientBoosting, LightGBM, and CatBoost. These branches represent complementary forms of randomized forests and gradient-boosted decision trees \cite{breiman2001random,chen2016xgboost,geurts2006extremely,friedman2001greedy,ke2017lightgbm,prokhorenkova2018catboost}.

For branch $m$, the output was

\begin{equation}
\mathbf{p}_{m}(\mathbf{x}_{i})
=
\left[
p_m(y=1\mid\mathbf{x}_{i}),
\ldots,
p_m(y=K\mid\mathbf{x}_{i})
\right]^{\mathsf{T}},
\label{eq:tree-branch-probabilities}
\end{equation}

with

\begin{equation}
\sum_{k=1}^{K}
p_m(y=k\mid\mathbf{x}_{i})
=
1.
\end{equation}

All branch probability arrays were aligned to the same 29-class ordering before fusion.

\subsubsection{Tree and Neural OOF Features}
\label{subsubsec:oof-feature-generation}

Stacked generalization requires higher-level models to be trained on predictions produced without fitting the corresponding base model directly on the same observation \cite{wolpert1992stacked,vanderlaan2007super}. The outer training set was therefore partitioned into $S=5$ stratified folds:

\begin{equation}
\mathcal{D}_{\mathrm{tr}}
=
\bigcup_{s=1}^{S}\mathcal{F}_{s},
\qquad
S=5.
\label{eq:oof-fold-partition}
\end{equation}

For tree branch $m$ and observation $i\in\mathcal{F}_{s}$, the OOF probability was

\begin{equation}
\mathbf{p}_{m,i}^{\mathrm{OOF}}
=
f_{m}^{(-s)}(\mathbf{x}_{i}),
\label{eq:tree-oof-probability}
\end{equation}

where $f_m^{(-s)}$ was fitted using
$\mathcal{D}_{\mathrm{tr}}\setminus\mathcal{F}_{s}$. After OOF generation, each tree branch was refitted on the complete outer training partition to obtain validation and test probabilities.

EcoFuseNet-V2 was processed similarly. A fold-specific neural model was trained without gradient updates from the held-out fold, after which the following held-out features were exported:

\begin{equation}
\left\{
\mathbf{p}_{i}^{\mathrm{Eco,OOF}},
\mathbf{e}_{i}^{\mathrm{OOF}},
\mathbf{g}_{i}^{\mathrm{OOF}},
\mathbf{d}_{i}^{\mathrm{OOF}}
\right\}.
\label{eq:neural-oof-features}
\end{equation}

The final EcoFuseNet-V2 model fitted on the complete outer training set generated the corresponding validation and test features.

This procedure prevented the meta-classifier from receiving neural outputs generated by a model trained directly on the same observation. However, the outer-training scaler was estimated before the inner folds were formed, and the held-out neural fold also supported checkpoint selection. The neural OOF implementation is therefore described as \emph{leakage-aware} rather than fully nested or strictly leakage-free.

\subsection{Hybrid Meta-Learning and Source Selection}
\label{subsec:hybrid-meta-learning}

\subsubsection{Hybrid Meta-Feature Matrix}
\label{subsubsec:hybrid-meta-feature-matrix}

For each observation, EcoTreeFuseNet-Plus concatenated the original standardized predictors, six tree-probability vectors, EcoFuseNet-V2 probabilities, ecological embedding, gate representation, and prototype distances:

\begin{equation}
\mathbf{Z}_{i}
=
\left[
\widehat{\mathbf{x}}_{i};
\mathbf{p}_{i}^{\mathrm{RF}};
\mathbf{p}_{i}^{\mathrm{XGB}};
\mathbf{p}_{i}^{\mathrm{ET}};
\mathbf{p}_{i}^{\mathrm{HGB}};
\mathbf{p}_{i}^{\mathrm{LGBM}};
\mathbf{p}_{i}^{\mathrm{CAT}};
\mathbf{p}_{i}^{\mathrm{Eco}};
\mathbf{e}_{i};
\mathbf{g}_{i};
\mathbf{d}_{i}
\right].
\label{eq:complete-meta-feature-vector}
\end{equation}

With all six tree branches available, the complete dimensionality was

\begin{equation}
\begin{aligned}
d_{\mathrm{meta}}
&=
8
+
6(29)
+
29
+
128
+
64
+
29
\\
&=
432.
\end{aligned}
\label{eq:meta-feature-dimension}
\end{equation}

The hybrid matrix was standardized using statistics estimated exclusively from the OOF training meta-features:

\begin{equation}
\widehat{Z}_{ij}
=
\frac{
Z_{ij}-\mu_{j}^{Z}
}{
\sigma_{j}^{Z}
},
\label{eq:meta-feature-standardization}
\end{equation}

and the same transformation was applied to validation and test meta-features.

\subsubsection{Candidate Meta-Classifiers}
\label{subsubsec:candidate-meta-classifiers}

Five meta-classifier families were compared:

\begin{enumerate}
    \item L2-regularized multinomial logistic regression;
    \item elastic-net multinomial logistic regression;
    \item HistGradientBoosting;
    \item ExtraTrees; and
    \item Random Forest.
\end{enumerate}

Elastic-net regularization permits simultaneous coefficient shrinkage and sparse feature selection \cite{zou2005regularization}. Each candidate was fitted on the OOF training meta-features and scored on the predefined validation partition. The selected meta-classifier was

\begin{equation}
h^{*}
=
\arg\max_{h\in\mathcal{H}}
\operatorname{MacroF1}
\left[
\mathbf{y}_{\mathrm{val}},
\arg\max_{k}
h_k\!\left(
\widehat{\mathbf{Z}}_{\mathrm{val}}
\right)
\right].
\label{eq:meta-classifier-selection}
\end{equation}

\subsubsection{Validation-Weighted Soft Voting}
\label{subsubsec:validation-soft-voting}

In parallel, a weighted soft-voting alternative combined the available tree and neural probability branches:

\begin{equation}
\mathbf{p}_{w}(\mathbf{x}_{i})
=
\sum_{m=1}^{M}
w_m
\mathbf{p}_{m}(\mathbf{x}_{i}),
\qquad
w_m\geq0,
\qquad
\sum_{m=1}^{M}w_m=1.
\label{eq:weighted-soft-voting}
\end{equation}

Candidate weights included uniform combinations, one-branch solutions, and randomly generated Dirichlet mixtures. The weight vector was selected using validation macro F1:

\begin{equation}
\mathbf{w}^{*}
=
\arg\max_{\mathbf{w}}
\operatorname{MacroF1}
\left(
\mathbf{y}_{\mathrm{val}},
\arg\max_{k}p_{w,k}
\right).
\label{eq:soft-vote-selection}
\end{equation}

Lower validation ECE was used to distinguish soft-voting candidates with equal macro F1.

The raw EcoTreeFuseNet-Plus probability source was then selected between the best meta-classifier and the optimized soft vote:

\begin{equation}
\mathbf{p}_{\mathrm{plus}}
=
\begin{cases}
\mathbf{p}_{\mathrm{meta}},
&
\operatorname{MacroF1}_{\mathrm{val}}
(\mathbf{p}_{\mathrm{meta}})
\geq
\operatorname{MacroF1}_{\mathrm{val}}
(\mathbf{p}_{w}),
\\[1mm]
\mathbf{p}_{w},
&
\text{otherwise}.
\end{cases}
\label{eq:base-plus-source-selection}
\end{equation}

The implemented base-source comparison used validation macro F1 directly; no additional ECE tie-breaking rule was applied at this final meta-versus-soft-vote selection stage.

\subsection{Calibration and Secondary Variants}
\label{subsec:calibration-secondary-variants}

\subsubsection{Temperature-Scaled Main Model}
\label{subsubsec:temperature-calibration}

After validation-based selection between the best meta-classifier and
the optimized soft-voting alternative, the resulting raw
EcoTreeFuseNet-Plus probability vector
$\mathbf{p}_{\mathrm{plus}}$ was calibrated using a single scalar
temperature \citep{guo2017calibration}. Calibration was performed on
probabilities rather than assuming direct access to model logits,
because the selected raw probability source could originate from
either meta-classification or probability-level soft voting.

For observation $i$ and class $k$, numerically stabilized
log-probabilities were defined as

\begin{equation}
\ell_{ik}
=
\log
\left[
\max
\left(
p_{\mathrm{plus},ik},
\epsilon
\right)
\right],
\label{eq:plus-log-probability}
\end{equation}

where $\epsilon$ is a small positive constant introduced to avoid
taking the logarithm of zero. The temperature-scaled probabilities
were then obtained as

\begin{equation}
p_{\mathrm{cal},ik}(T)
=
\frac{
\exp\!\left(\ell_{ik}/T\right)
}{
\displaystyle
\sum_{r=1}^{K}
\exp\!\left(\ell_{ir}/T\right)
},
\qquad
T>0.
\label{eq:temperature-scaling}
\end{equation}

Equivalently, the probability transformation can be written compactly
as

\begin{equation}
\mathbf{p}_{\mathrm{cal},i}(T)
=
\operatorname{softmax}
\left(
\frac{
\log \mathbf{p}_{\mathrm{plus},i}
}{T}
\right),
\label{eq:temperature-scaling-compact}
\end{equation}

with numerical clipping applied before the logarithm as defined in
Eq.~\ref{eq:plus-log-probability}.

The temperature was selected exclusively from the validation
partition by minimizing negative log-likelihood:

\begin{equation}
T^{*}
=
\arg\min_{T\in\mathcal{G}_{T}}
\operatorname{NLL}
\left(
\mathbf{y}_{\mathrm{val}},
\mathbf{p}_{\mathrm{cal}}(T)
\right),
\label{eq:temperature-selection}
\end{equation}

where the implemented calibration grid was restricted to

\begin{equation}
\mathcal{G}_{T}\subseteq[0.5,5.0].
\label{eq:temperature-grid}
\end{equation}

Within this prespecified validation grid, the selected value was
$T^{*}=0.50$. Because the selected value coincided with the lower
boundary of the implemented search interval, it is interpreted as the
validation-NLL optimum within the evaluated grid rather than as
evidence of an unconstrained global optimum. The resulting
$\mathbf{p}_{\mathrm{cal}}(T^{*})$ defined the principal proposed
model, Calibrated EcoTreeFuseNet-Plus. Temperature scaling was applied
only after the raw probability source had been fixed from validation
data, and the test labels were not involved in temperature selection.

\subsubsection{Bias-Tuned Variant}
\label{subsubsec:bias-tuned-variant}

A secondary class-bias variant was constructed to investigate whether
validation-based class-specific probability adjustment could improve
macro-averaged discrimination. Class-specific offsets were introduced
into the calibrated log-probabilities:

\begin{equation}
p_{\mathrm{bias},ik}
=
\operatorname{softmax}_{k}
\left(
\log p_{\mathrm{cal},ik}
+
b_k
\right),
\label{eq:bias-tuned-probability}
\end{equation}

where $b_k$ denotes the offset associated with fine class $k$. The
bias vector was optimized using validation-only greedy coordinate
search:

\begin{equation}
\mathbf{b}^{*}
=
\arg\max_{\mathbf{b}}
\operatorname{MacroF1}
\left[
\mathbf{y}_{\mathrm{val}},
\arg\max_k
\left(
\log p_{\mathrm{cal},ik}+b_k
\right)
\right],
\label{eq:bias-selection}
\end{equation}

subject to

\begin{equation}
-1.25
\leq
b_k
\leq
1.25,
\qquad
k=1,\ldots,K.
\label{eq:bias-range}
\end{equation}

The bias-tuned model was treated as a secondary
minority-class-sensitive variant. It was not used to define the
principal proposed model or the main statistical reference.

\subsubsection{Hierarchy-Aware Variant}
\label{subsubsec:hierarchy-aware-variant}

A second secondary variant incorporated consistency between the
29 fine classes and their five broader ecological categories. Let
$\phi(k)$ denote the coarse ecological category associated with fine
class $k$. For observation $i$, the probability mass associated with
coarse category $c$ was calculated from the bias-tuned fine-class
probabilities as

\begin{equation}
S_{ic}
=
\sum_{k:\phi(k)=c}
p_{\mathrm{bias},ik}.
\label{eq:coarse-probability-mass}
\end{equation}

The hierarchy-adjusted fine-class probabilities were then defined as

\begin{equation}
p_{\mathrm{hier},ik}(\alpha)
=
\frac{
p_{\mathrm{bias},ik}
S_{i,\phi(k)}^{\alpha}
}{
\displaystyle
\sum_{r=1}^{K}
p_{\mathrm{bias},ir}
S_{i,\phi(r)}^{\alpha}
},
\label{eq:hierarchical-probability}
\end{equation}

where $\alpha\geq0$ controls the strength of the coarse-category
adjustment. Setting $\alpha=0$ recovers the bias-tuned fine-class
probabilities without an additional hierarchy contribution.

The hierarchy strength was selected exclusively on the validation
partition:

\begin{equation}
\alpha^{*}
=
\arg\max_{\alpha\in\mathcal{G}_{\alpha}}
\operatorname{MacroF1}
\left[
\mathbf{y}_{\mathrm{val}},
\arg\max_k
p_{\mathrm{hier},ik}(\alpha)
\right],
\label{eq:hierarchy-alpha-selection}
\end{equation}

using the candidate grid

\begin{equation}
\mathcal{G}_{\alpha}
=
\left\{
0,\,
0.15,\,
0.25,\,
0.5,\,
0.75,\,
1,\,
1.5,\,
2
\right\}.
\label{eq:hierarchy-grid}
\end{equation}

When two hierarchy candidates produced identical validation macro F1,
the candidate with the lower validation ECE was preferred. The
hierarchy-aware model was retained only as a secondary
ecological-consistency analysis and did not replace Calibrated
EcoTreeFuseNet-Plus as the principal proposed model.

Overall, the four reported probability outputs therefore represent
distinct roles within the experimental design:
$\mathbf{p}_{\mathrm{plus}}$ is the validation-selected uncalibrated
fusion output,
$\mathbf{p}_{\mathrm{cal}}(T^{*})$ defines the principal calibrated
model,
$\mathbf{p}_{\mathrm{bias}}(\mathbf{b}^{*})$ represents the
class-bias-adjusted secondary variant, and
$\mathbf{p}_{\mathrm{hier}}(\alpha^{*})$ represents the
hierarchy-aware secondary variant. All calibration and secondary
probability transformations were selected using validation data only
before evaluation on the held-out test partition.

\subsection{Model Definitions and Fixed Configuration}
\label{subsec:model-definitions-configuration}

The four reported framework variants are summarized in Table~\ref{tab:framework-variants}.

\begin{table}[htbp]
\centering
\caption{Definitions and reporting roles of the EcoTreeFuseNet-Plus variants.}
\label{tab:framework-variants}
\begin{tabularx}{\textwidth}{
>{\raggedright\arraybackslash}p{0.31\textwidth}
>{\raggedright\arraybackslash}p{0.25\textwidth}
>{\raggedright\arraybackslash}X}
\toprule
\textbf{Variant} &
\textbf{Probability output} &
\textbf{Role in the study} \\
\midrule

EcoTreeFuseNet-Plus &
$\mathbf{p}_{\mathrm{plus}}$ &
Validation-selected uncalibrated meta-classifier or soft-voting output \\

Calibrated EcoTreeFuseNet-Plus &
$\mathbf{p}_{\mathrm{cal}}(T^{*})$ &
Principal proposed model and main statistical reference \\

Bias-Tuned EcoTreeFuseNet-Plus &
$\mathbf{p}_{\mathrm{bias}}(\mathbf{b}^{*})$ &
Validation-tuned secondary variant emphasizing macro F1 \\

Hierarchy-Aware EcoTreeFuseNet-Plus &
$\mathbf{p}_{\mathrm{hier}}(\alpha^{*})$ &
Secondary variant incorporating fine-to-coarse ecological consistency \\
\bottomrule
\end{tabularx}
\end{table}

The fixed framework-level constants are reported in Table~\ref{tab:framework-hyperparameters}. Estimator-specific settings for the tree and meta-classifier candidates were held fixed within the executable configuration and were applied consistently across all corresponding comparisons.

\begin{table}[htbp]
\centering
\caption{Fixed architecture, optimization, and fusion constants.}
\label{tab:framework-hyperparameters}

\small
\renewcommand{\arraystretch}{1.10}
\setlength{\tabcolsep}{4pt}

\begin{tabularx}{\textwidth}{
>{\raggedright\arraybackslash}p{0.29\textwidth}
>{\centering\arraybackslash}p{0.23\textwidth}
>{\raggedright\arraybackslash}X
}
\toprule
\textbf{Setting} &
\textbf{Value} &
\textbf{Purpose} \\
\midrule

Fine classes, $K$
& 29
& Primary prediction categories \\

Coarse groups, $C$
& 5
& Auxiliary ecological hierarchy \\

Branch width, $d_f$
& 64
& Terrain, CHM, spectral, and gate dimensions \\

Embedding size, $d_e$
& 128
& Ecological latent representation \\

Initial $\alpha_{\mathrm{CHM}}$
& 0.05
& Restrained initial canopy contribution \\

Minimum prototype temperature
& 0.1
& Ensures strictly positive temperature \\

Label smoothing, $\varepsilon$
& 0.03
& Smoothed fine-class targets \\

Full $\lambda_p$
& 0.03
& Prototype-contrastive weight \\

Full $\lambda_c$
& 0.01
& Coarse-consistency weight \\

Full $\lambda_s$
& 0.001
& Prototype-separation weight \\

OOF $\lambda_p$
& 0.01
& Fold-specific prototype weight \\

OOF $\lambda_c$
& 0.005
& Fold-specific coarse weight \\

OOF $\lambda_s$
& 0.0005
& Fold-specific separation weight \\

Prototype margin, $m$
& 1.0
& Bounded fine-prototype separation \\

Gradient-norm limit
& 1.0
& Neural optimization stability \\

OOF folds, $S$
& 5
& Tree and neural meta-feature generation \\

Temperature interval
& $[0.5,5.0]$
& Validation-NLL calibration search \\

Class-bias interval
& $[-1.25,1.25]$
& Validation-only bias search \\

Hierarchy grid
&
\shortstack[c]{
$\{0,\,0.15,\,0.25,\,0.5,$\\
$0.75,\,1,\,1.5,\,2\}$
}
& Validation-only hierarchy selection \\

\bottomrule
\end{tabularx}

\normalsize
\end{table}

\subsection{Complete Training and Inference Algorithm}
\label{subsec:complete-framework-algorithm}

The complete procedure is summarized in Algorithm~\ref{alg:ecotreefusenet-plus}.

\begin{algorithm}[htbp]
\caption{Training and inference procedure for Calibrated EcoTreeFuseNet-Plus.}
\label{alg:ecotreefusenet-plus}
\begin{algorithmic}[1]
\Require Training, validation, and test partitions; fine labels $y$; coarse labels $c$; mapping $\phi$
\Ensure Calibrated test probabilities and predicted fine classes

\State Apply the CHM transformation and training-derived standardization
\State Partition the predictors into terrain, canopy, and spectral groups
\State Compute class weights from the outer training labels
\State Train EcoFuseNet-V2 using cross-entropy warm-up
\State Initialize fine and coarse prototypes from training embeddings
\State Continue EcoFuseNet-V2 training using weak prototype regularization
\State Configure the six tree-probability branches

\For{$s=1,\ldots,5$}
    \State Generate held-out tree probabilities for fold $\mathcal{F}_{s}$
    \State Train fold-specific EcoFuseNet-V2
    \State Export held-out neural probabilities, embeddings, gates, and distances
\EndFor

\State Refit branch models on the complete outer training set
\State Generate aligned validation and test branch outputs
\State Construct the 432-dimensional training, validation, and test meta-feature matrices
\State Fit the meta-feature scaler using OOF training features only
\State Train candidate meta-classifiers and select one using validation macro F1
\State Optimize validation-weighted soft voting
\State Select the raw Plus source using validation macro F1
\State Select $T^{*}$ by minimizing validation NLL
\State Apply $T^{*}$ to the corresponding test probabilities
\State Optionally generate the validation-selected bias-tuned and hierarchy-aware variants
\State Predict
$\widehat{y}_{i}
=
\arg\max_{k}p_{\mathrm{cal},ik}(T^{*})$
\end{algorithmic}
\end{algorithm}

All meta-learning, soft-voting, calibration, bias, and hierarchy decisions were based on the same predefined validation partition. The test labels were accessed only after the complete configuration and probability transformations had been fixed.

\subsection{Computational and Reproducibility Details}
\label{subsec:framework-reproducibility}

The complete framework was implemented in Python. PyTorch was used for
EcoFuseNet-V2 and neural optimization, while scikit-learn-compatible
tools supported preprocessing, conventional ensembles,
meta-classifiers, and evaluation
\citep{paszke2019pytorch,pedregosa2011scikit}. XGBoost, LightGBM, and
CatBoost were accessed through their corresponding Python interfaces.
AdamW was used for neural optimization. The principal architecture,
loss, out-of-fold, calibration, and secondary-variant settings are
reported in Table~\ref{tab:framework-hyperparameters}.

All reported experiments were executed in a CPU environment without
GPU acceleration. Random states were propagated, wherever supported,
to data partitioning, conventional learners, neural initialization,
mini-batch ordering, out-of-fold generation, candidate soft-voting
weight generation, and repeated-split evaluation. The repeated-seed
protocol itself is defined in
Section~\ref{subsec:repeated-seed-protocol}.

For each run, the implementation retained partition metadata,
validation-selected neural checkpoints, refitted branch outputs,
out-of-fold probability arrays, EcoFuseNet-V2 representations,
meta-feature scalers, selected meta-classifiers, soft-voting weights,
selected raw probability sources, calibration parameters,
secondary-variant parameters, final test probabilities, predicted
labels, and statistical outputs. These saved artifacts allowed the
reported metrics, calibration analyses, figures, and paired
statistical comparisons to be regenerated from fixed predictions
without retraining the complete pipeline.
\clearpage

\section{Experimental Results}
\label{sec:experimental-results}

\subsection{Experimental Configuration}
\label{subsec:results-configuration}

The primary evaluation used the spatially informed seed-42 partition
described in Section~\ref{sec:experimental-design}. All baseline,
ablation, neural, and hybrid models were evaluated using the same
training, validation, and test observations. The complete seed-42
benchmark suite, including the baseline, ablation, hybrid, and
diagnostic experiments, was executed on a CPU and required
20.56~min. The principal configuration is summarized in
Table~\ref{tab:results-configuration}.

\begin{table}[!ht]
\centering
\caption{Experimental configuration for the primary seed-42 evaluation.}
\label{tab:results-configuration}
\begin{tabular}{lr}
\toprule
\textbf{Setting} & \textbf{Value} \\
\midrule
Total observations & 1,833 \\
Training observations & 1,273 \\
Validation observations & 190 \\
Test observations & 370 \\
Fine classes & 29 \\
Predictors & 8 \\
Primary random seed & 42 \\
Out-of-fold stacking folds & 5 \\
Bootstrap resamples & 1,000 \\
Execution device & CPU \\
Complete seed-42 benchmark runtime & 20.56 min \\
\bottomrule
\end{tabular}
\end{table}

\subsection{Overall Predictive Performance}
\label{subsec:overall-performance}

Table~\ref{tab:results-overall-performance} compares the standalone
tree ensembles, EcoFuseNet-V2, the legacy hybrid model, and the
EcoTreeFuseNet-Plus variants. Calibrated EcoTreeFuseNet-Plus
correctly classified 296 of the 370 test observations, yielding an
accuracy of 0.8000, a macro F1-score of 0.7768, a balanced accuracy
of 0.7903, and an MCC of 0.7903. Its weighted F1-score and Cohen's
kappa were 0.8010 and 0.7900, respectively.

ExtraTrees was the strongest standalone baseline, with an accuracy
of 0.7865, a macro F1-score of 0.7712, a balanced accuracy of
0.7618, and an MCC of 0.7759. The proposed model therefore exceeded
ExtraTrees by 0.0055 in macro F1. Legacy EcoTreeFuseNet obtained a
macro F1-score of 0.7549, whereas standalone EcoFuseNet-V2 obtained
0.5639.

The raw and calibrated EcoTreeFuseNet-Plus variants produced
identical class labels and consequently identical discrimination
metrics. The bias-tuned and hierarchy-aware variants slightly reduced
ECE to 0.0597 but also reduced accuracy and macro F1. The selected
hierarchy coefficient was $\alpha=0$, causing the hierarchy-aware and
bias-tuned variants to produce identical predictions and metrics.

\begin{table}[!htbp]
\centering
\caption{Test-set discrimination and probability-reliability
performance under the primary seed-42 split. Lower values are
preferable for ECE, Brier score, and NLL.}
\label{tab:results-overall-performance}
\small
\resizebox{\linewidth}{!}{%
\begin{tabular}{lrrrrrrr}
\toprule
\textbf{Model} &
\textbf{Accuracy} &
\textbf{Macro F1} &
\textbf{Bal. Acc.} &
\textbf{MCC} &
\textbf{ECE} &
\textbf{Brier} &
\textbf{NLL} \\
\midrule
Random Forest
& 0.7676 & 0.7500 & 0.7349 & 0.7559 & 0.2386 & 0.4098 & 0.8943 \\

XGBoost
& 0.7730 & 0.7448 & 0.7449 & 0.7618 & 0.0808 & 0.3379 & 0.7899 \\

ExtraTrees
& 0.7865 & 0.7712 & 0.7618 & 0.7759 & 0.2910 & 0.4187 & 0.9346 \\

HistGradientBoosting
& 0.7243 & 0.7170 & 0.7126 & 0.7106 & 0.0700 & 0.3780 & 0.9024 \\

LightGBM
& 0.7703 & 0.7407 & 0.7517 & 0.7589 & 0.0823 & 0.3464 & 0.8303 \\

CatBoost
& 0.7486 & 0.7398 & 0.7561 & 0.7373 & 0.1684 & 0.3964 & 0.8685 \\

EcoFuseNet-V2
& 0.6054 & 0.5639 & 0.5720 & 0.5869 & 0.0702 & 0.5302 & 1.1421 \\

Legacy EcoTreeFuseNet
& 0.7811 & 0.7549 & 0.7687 & 0.7704 & 0.1984 & 0.3766 & 0.9239 \\

EcoTreeFuseNet-Plus (raw)
& 0.8000 & 0.7768 & 0.7903 & 0.7903 & 0.3866 & 0.4839 & 1.2926 \\

\textbf{Calibrated EcoTreeFuseNet-Plus}
& 0.8000 & 0.7768 & 0.7903 & 0.7903 & 0.0651 & 0.3108 & 0.8179 \\

Bias-Tuned EcoTreeFuseNet-Plus
& 0.7811 & 0.7653 & 0.7772 & 0.7708 & 0.0597 & 0.3367 & 0.9072 \\

Hierarchy-Aware EcoTreeFuseNet-Plus
& 0.7811 & 0.7653 & 0.7772 & 0.7708 & 0.0597 & 0.3367 & 0.9072 \\
\bottomrule
\end{tabular}}
\end{table}

\FloatBarrier

\subsection{Fusion Selection and Calibration Performance}
\label{subsec:fusion-calibration-results}

All fusion and calibration decisions were based exclusively on
validation-set performance. Elastic-net multinomial logistic
regression with $C=0.003$ achieved the highest validation macro
F1-score of 0.7037 and was selected as the meta-classifier. The best
validation-weighted soft-voting solution achieved a macro F1-score
of 0.6979 and was therefore not selected as the final base-probability
source.

The optimized soft-voting solution assigned 85.88\% of its weight
to ExtraTrees and 8.88\% to CatBoost. The remaining branches received
substantially smaller weights. The displayed weights sum to 99.99\%
because of rounding.

Temperature scaling selected $T=0.50$ by minimizing validation NLL.
The resulting validation NLL and ECE were 0.8764 and 0.0868,
respectively. The complete validation-based selection results are
reported in Table~\ref{tab:fusion-selection}.

\begin{table}[!ht]
\centering
\caption{Validation-based selection of the meta-classifier,
soft-voting alternative, and calibration temperature.}
\label{tab:fusion-selection}

\small
\setlength{\tabcolsep}{4pt}
\renewcommand{\arraystretch}{1.08}

\begin{tabular*}{0.78\linewidth}{
@{\extracolsep{\fill}}
l
r
c
@{}
}
\toprule

\multicolumn{3}{l}{
\textit{Panel A: Candidate meta-classifiers}
} \\

\midrule

\textbf{Candidate} &
\textbf{Val. Macro F1} &
\textbf{Selected} \\

\midrule

L2 logistic, $C=0.001$
& 0.6837
& No \\

L2 logistic, $C=0.003$
& 0.6895
& No \\

L2 logistic, $C=0.010$
& 0.6745
& No \\

L2 logistic, $C=0.030$
& 0.6723
& No \\

L2 logistic, $C=0.100$
& 0.6635
& No \\

Elastic-net logistic, $C=0.003$
& 0.7037
& Yes \\

Elastic-net logistic, $C=0.010$
& 0.6950
& No \\

Elastic-net logistic, $C=0.030$
& 0.6580
& No \\

HistGradientBoosting
& 0.6418
& No \\

ExtraTrees
& 0.6586
& No \\

Random Forest
& 0.6559
& No \\

\bottomrule
\end{tabular*}

\vspace{8pt}

\begin{tabular*}{\linewidth}{
@{\extracolsep{\fill}}
l
r
l
r
@{}
}
\toprule

\multicolumn{4}{l}{
\textit{Panel B: Soft-voting weights and final selection}
} \\

\midrule

\textbf{Branch} &
\textbf{Weight} &
\textbf{Selection quantity} &
\textbf{Value} \\

\midrule

Random Forest
& 0.49\%
& Soft-vote val. Macro F1
& 0.6979 \\

XGBoost
& 1.97\%
& Meta-model val. Macro F1
& 0.7037 \\

ExtraTrees
& 85.88\%
& Selected base source
& Meta-classifier \\

HistGradientBoosting
& 0.00\%
& Temperature, $T$
& 0.50 \\

LightGBM
& 1.03\%
& Validation NLL
& 0.8764 \\

CatBoost
& 8.88\%
& Validation ECE
& 0.0868 \\

EcoFuseNet-V2
& 1.74\%
& {}
& {} \\

\bottomrule
\end{tabular*}

\normalsize
\end{table}

Temperature scaling substantially improved probability reliability
without altering the predicted labels. ECE decreased from 0.3866 to
0.0651, corresponding to an 83.16\% reduction. The Brier score
decreased by 35.78\%, from 0.4839 to 0.3108, while NLL decreased by
36.72\%, from 1.2926 to 0.8179. The discrimination metrics remained
unchanged, as summarized in Table~\ref{tab:results-calibration-effect}.

\begin{table}[!htbp]
\centering
\caption{Effect of temperature scaling on EcoTreeFuseNet-Plus.
Predicted class labels were unchanged.}
\label{tab:results-calibration-effect}
\begin{tabular}{lrrr}
\toprule
\textbf{Metric} &
\textbf{Raw Plus} &
\textbf{Calibrated Plus} &
\textbf{Relative change} \\
\midrule
Accuracy & 0.8000 & 0.8000 & Unchanged \\
Macro F1 & 0.7768 & 0.7768 & Unchanged \\
Balanced accuracy & 0.7903 & 0.7903 & Unchanged \\
MCC & 0.7903 & 0.7903 & Unchanged \\
ECE & 0.3866 & 0.0651 & $-83.16$\% \\
Brier score & 0.4839 & 0.3108 & $-35.78$\% \\
NLL & 1.2926 & 0.8179 & $-36.72$\% \\
\bottomrule
\end{tabular}
\end{table}

\begin{figure}[!htb]
\centering
\includegraphics[width=0.98\linewidth]
{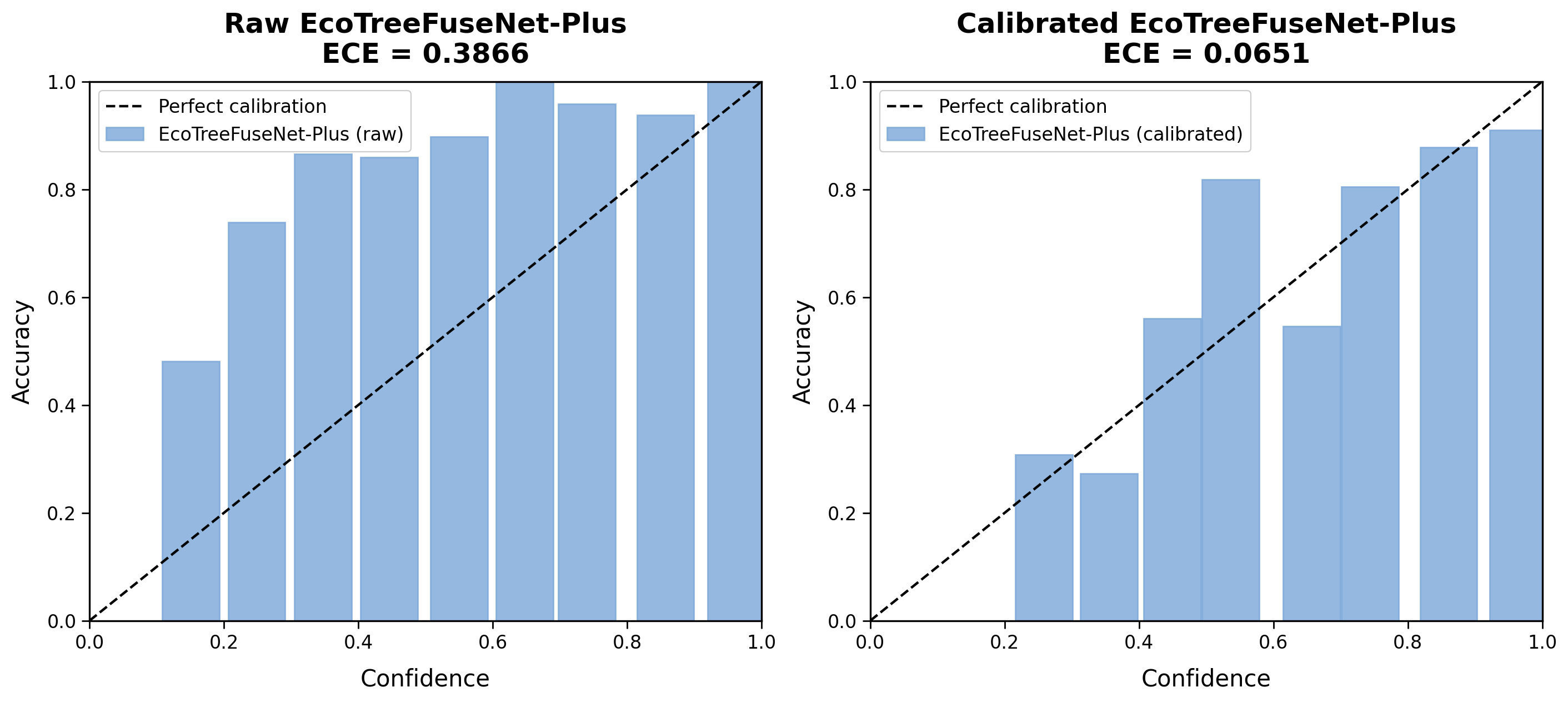}
\caption{Reliability diagrams of EcoTreeFuseNet-Plus before and after
temperature scaling. Calibration reduced the expected calibration
error from 0.3866 to 0.0651 without altering the predicted class
labels.}
\label{fig:results-reliability}
\end{figure}

Figure~\ref{fig:results-reliability} compares the raw and calibrated
reliability profiles. The raw model was predominantly underconfident,
with empirical accuracy exceeding mean confidence in most occupied
bins. Following temperature scaling, the confidence levels were
substantially closer to the corresponding empirical accuracies.

\FloatBarrier

\subsection{Statistical Robustness and Ablation Analysis}
\label{subsec:statistical-ablation-results}

Bootstrap resampling produced a 95\% confidence interval of
$[0.7059,0.8152]$ for macro F1 and $[0.7323,0.8379]$ for balanced
accuracy. Relative to ExtraTrees, the proposed model obtained a
macro-F1 difference of 0.0055, with a paired-bootstrap interval of
$[-0.0362,0.0499]$ and $p=0.733$. The balanced-accuracy difference
and McNemar comparison against ExtraTrees were also non-significant.

For macro F1, statistically supported differences were observed
against HistGradientBoosting, LightGBM, and EcoFuseNet-V2, whereas
the differences relative to Random Forest, XGBoost, ExtraTrees,
CatBoost, and Legacy EcoTreeFuseNet were not significant. Balanced
accuracy was significantly higher than that of Random Forest,
XGBoost, HistGradientBoosting, LightGBM, and EcoFuseNet-V2.
McNemar's test identified significant differences in paired
correctness relative to HistGradientBoosting, LightGBM, CatBoost,
and EcoFuseNet-V2. Because the reported probabilities were not
adjusted for multiple comparisons, they are interpreted as
model-specific evidence rather than simultaneous family-wise tests.

\begin{table}[!htbp]
\centering
\caption{Bootstrap uncertainty and paired comparison of Calibrated
EcoTreeFuseNet-Plus with the principal competing models. Differences
are calculated as proposed model minus comparator. Paired-bootstrap
$p$-values are two-sided and unadjusted for multiplicity.}
\label{tab:results-statistical-comparison}
\small

\begin{tabular}{ll}
\toprule
Macro F1, observed [95\% CI]
& 0.7768 [0.7059, 0.8152] \\
Balanced accuracy, observed [95\% CI]
& 0.7903 [0.7323, 0.8379] \\
\bottomrule
\end{tabular}

\vspace{5pt}

\resizebox{\linewidth}{!}{%
\begin{tabular}{lrrrrr}
\toprule
\textbf{Comparator} &
\textbf{$\Delta$Macro F1 [95\% CI]} &
\textbf{$p$} &
\textbf{$\Delta$Bal. Acc. [95\% CI]} &
\textbf{$p$} &
\textbf{McNemar $p$} \\
\midrule
Random Forest
& 0.0268 [$-0.0164$, 0.0732]
& 0.230
& 0.0553 [0.0163, 0.0929]
& 0.010
& 0.074 \\

XGBoost
& 0.0320 [$-0.0039$, 0.0721]
& 0.072
& 0.0454 [0.0135, 0.0785]
& 0.008
& 0.052 \\

ExtraTrees
& 0.0055 [$-0.0362$, 0.0499]
& 0.733
& 0.0285 [$-0.0116$, 0.0668]
& 0.156
& 0.542 \\

HistGradientBoosting
& 0.0597 [0.0166, 0.1109]
& 0.006
& 0.0777 [0.0335, 0.1225]
& 0.002
& $<0.001$ \\

LightGBM
& 0.0361 [0.0090, 0.0716]
& 0.014
& 0.0385 [0.0105, 0.0712]
& 0.008
& 0.027 \\

CatBoost
& 0.0369 [$-0.0047$, 0.0775]
& 0.086
& 0.0341 [$-0.0072$, 0.0721]
& 0.110
& 0.005 \\

EcoFuseNet-V2
& 0.2129 [0.1434, 0.2790]
& 0.002
& 0.2182 [0.1411, 0.2829]
& 0.002
& $<0.001$ \\

Legacy EcoTreeFuseNet
& 0.0219 [$-0.0160$, 0.0600]
& 0.240
& 0.0216 [$-0.0130$, 0.0569]
& 0.240
& 0.210 \\
\bottomrule
\end{tabular}}
\end{table}

The architectural and modality ablations are summarized in
Table~\ref{tab:results-ablation}. EcoFuseNet-V2 achieved a macro
F1-score of 0.5639. The No-Gate MLP and No-CHM-gate variants
achieved macro F1-scores of 0.4412 and 0.4410, respectively. The
flat-loss variant achieved 0.5427, while the legacy prototype model
achieved 0.4835.

The LiDAR-only and spectral-only configurations produced macro
F1-scores of 0.1930 and 0.2224, respectively. All neural-only and
modality-restricted variants remained below the complete calibrated
tree--neural framework.

\begin{table}[!ht]
\centering
\caption{Architectural and modality ablation results under the
common seed-42 protocol.}
\label{tab:results-ablation}

\small
\setlength{\tabcolsep}{5pt}
\renewcommand{\arraystretch}{1.05}

\begin{tabular*}{\linewidth}{
@{\extracolsep{\fill}}
l
r
r
r
r
r
@{}
}
\toprule
\textbf{Model} &
\textbf{Accuracy} &
\textbf{Macro F1} &
\textbf{Bal. Acc.} &
\textbf{MCC} &
\textbf{ECE} \\
\midrule

EcoFuseNet-V2
& 0.6054
& 0.5639
& 0.5720
& 0.5869
& 0.0702 \\

Legacy prototype model
& 0.4541
& 0.4835
& 0.5242
& 0.4347
& 0.3364 \\

No-Gate MLP
& 0.4514
& 0.4412
& 0.4791
& 0.4337
& 0.3587 \\

Flat-loss model
& 0.6081
& 0.5427
& 0.5373
& 0.5885
& 0.0668 \\

LiDAR-only model
& 0.2216
& 0.1930
& 0.2600
& 0.1970
& 0.5016 \\

Spectral-only model
& 0.2297
& 0.2224
& 0.2605
& 0.2041
& 0.3097 \\

No-CHM-gate model
& 0.4676
& 0.4410
& 0.4827
& 0.4479
& 0.3744 \\

Calibrated EcoTreeFuseNet-Plus
& 0.8000
& 0.7768
& 0.7903
& 0.7903
& 0.0651 \\

\bottomrule
\end{tabular*}

\normalsize
\end{table}

\FloatBarrier

\subsection{Class-Wise and Error Analysis}
\label{subsec:classwise-error-results}

Class-wise evaluation revealed substantial variation in performance
across the 29 fine ecological classes. IrrigLawnPasture achieved the
highest F1-score of 0.9863, followed by Water at 0.9630 and Wyethia
at 0.9091. GravelRoadTrail, Snow, and WetAlpine each achieved an
F1-score of 0.8889. These results indicate that several classes were
identified consistently despite the fine-grained and imbalanced
nature of the classification problem.

The lowest F1-score was observed for DrySubalpTill at 0.5143.
SandstoneOutcrop and SandstoneTalus each achieved an F1-score of
0.5714, whereas GraniteTalus and Paved each obtained 0.6154.
Performance estimates for low-support classes were interpreted
cautiously. For example, BareSoil and Frasera each achieved an
F1-score of 0.8571, but both classes contained only four observations
in the seed-42 test partition. Consequently, apparently high
class-specific scores for such categories should not be interpreted
as equally stable as those obtained for better represented classes.

\begin{figure}[!htb]
\centering
\includegraphics[
    width=0.90\linewidth,
    height=0.58\textheight,
    keepaspectratio
]{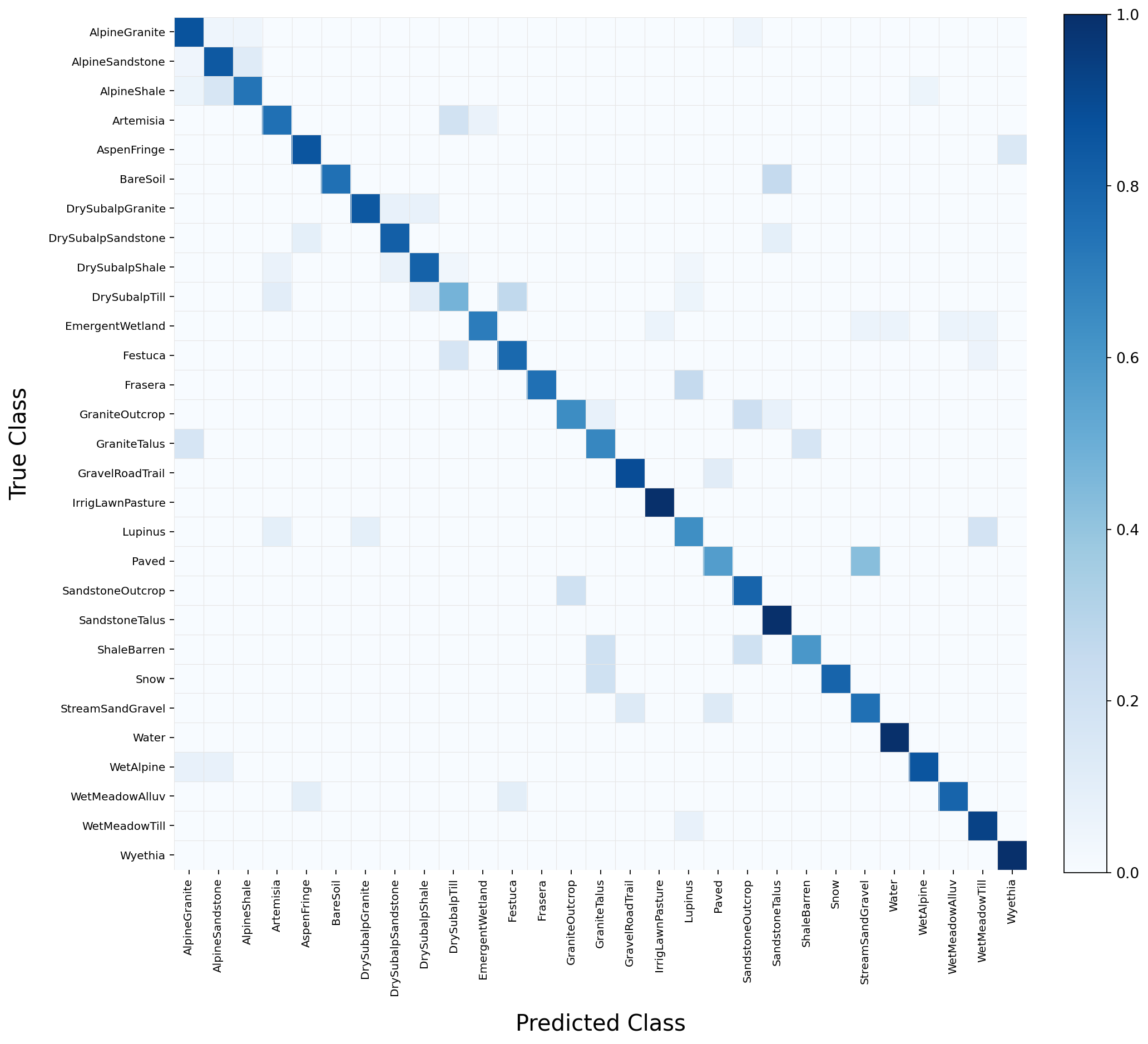}
\caption{Row-normalized confusion matrix of Calibrated
EcoTreeFuseNet-Plus on the seed-42 test partition ($N=370$).
The model correctly classified 296 observations and misclassified
74 observations.}
\label{fig:results-confusion}
\end{figure}

The row-normalized confusion matrix is shown in
Figure~\ref{fig:results-confusion}. The largest bidirectional
confusion occurred between DrySubalpTill and Festuca, with eight
combined errors. AlpineSandstone and AlpineShale produced six mutual
errors, while Artemisia and DrySubalpTill produced five. Other
prominent confusion pairs were Paved and StreamSandGravel, and
GraniteOutcrop and SandstoneOutcrop, each with four combined errors.
These error patterns indicate that the remaining classification
difficulty was concentrated primarily among fine classes with
overlapping ecological or surface characteristics rather than being
distributed uniformly across the target taxonomy.

\FloatBarrier

\subsection{Meta-Feature and Neural-Branch Diagnostics}
\label{subsec:model-diagnostic-results}

\begin{figure}[!htb]
\centering
\includegraphics[
    width=0.88\linewidth,
    height=0.48\textheight,
    keepaspectratio
]{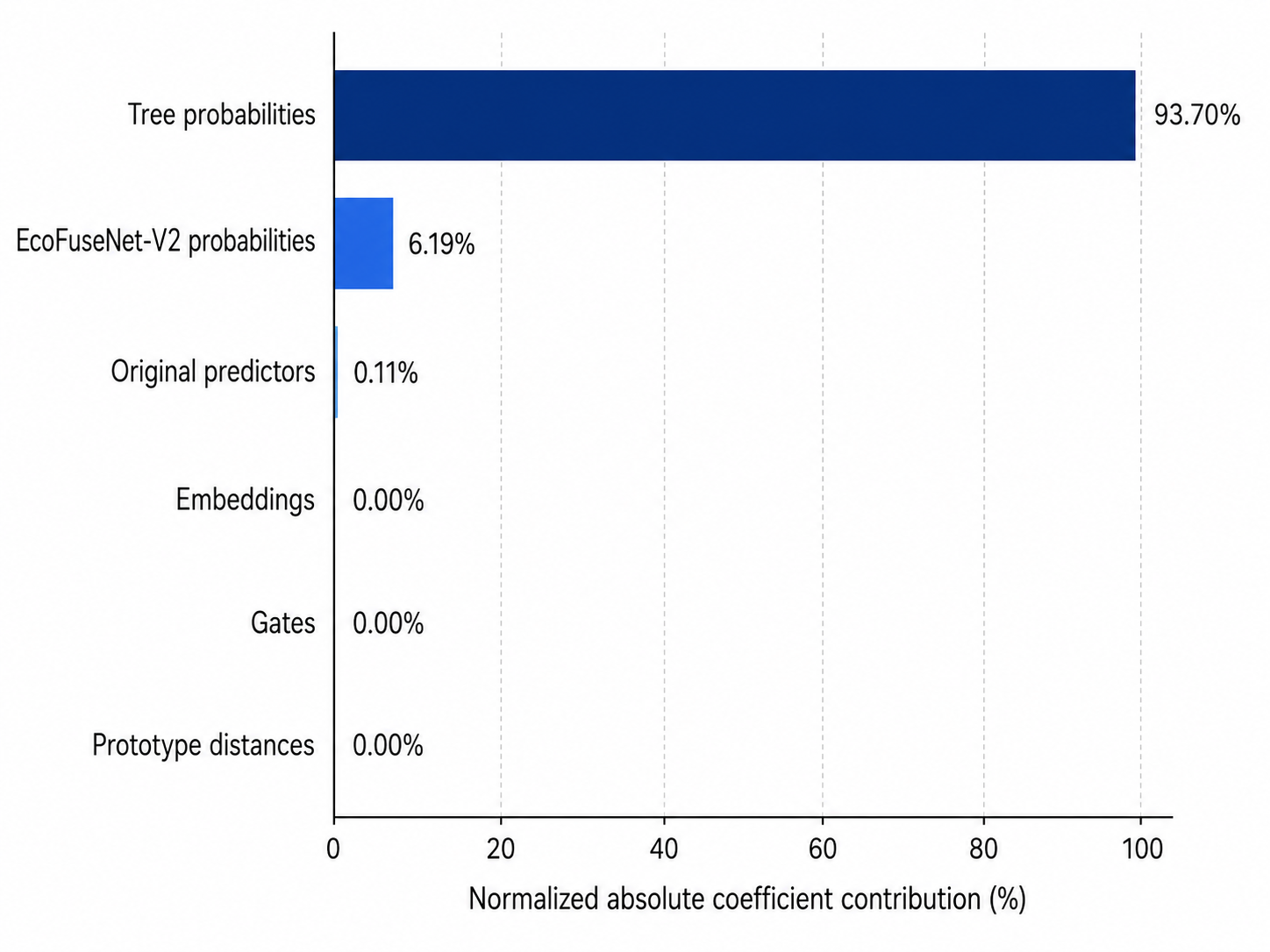}
\caption{Normalized absolute coefficient contributions of the
meta-feature families retained by the selected elastic-net
meta-classifier. Tree-probability features accounted for 93.70\%
of the total contribution, followed by EcoFuseNet-V2 class
probabilities at 6.19\% and the original predictors at 0.11\%.
The embedding, gate, and prototype-distance feature families
received zero coefficient contribution.}
\label{fig:results-meta-feature}
\end{figure}

The grouped contribution of the selected elastic-net meta-classifier
is shown in Figure~\ref{fig:results-meta-feature}. Tree-probability
features accounted for 93.70\% of the normalized absolute coefficient
contribution, while EcoFuseNet-V2 class probabilities contributed
6.19\% and the original predictors contributed 0.11\%. The remaining
latent neural feature families---ecological embeddings, gate outputs,
and prototype-distance features---received zero coefficient
contribution in the selected sparse meta-classifier.

Among the eight original predictors, only NDVI was retained, with an
absolute elastic-net coefficient magnitude of 0.000656. These results
indicate that the final meta-classifier was driven predominantly by
tree-derived class probabilities, with a smaller complementary
contribution from EcoFuseNet-V2 probabilities. The zero coefficients
assigned to the latent neural feature families indicate exclusion
under the selected regularization setting and should not be
interpreted as evidence that those representations contain no
information.

The class-level gate analysis did not provide evidence of the
predefined ecological relationship with canopy height. The Spearman
association between class-level mean CHM and mean gate value was weak
and non-significant ($\rho=0.124$, $p=0.521$). The association
between class-level positive-CHM proportion and mean gate value was
similarly weak and non-significant ($\rho=0.132$, $p=0.496$).
Accordingly, the predefined ecological gate-validity criterion was
not satisfied.

The prototype diagnostic yielded a macro compactness ratio of 0.6906.
The mean within-class distance to the corresponding learned prototype
was 6.3266, while the mean nearest inter-prototype distance was
4.3626. These quantities are reported as exploratory diagnostics of
the EcoFuseNet-V2 representation geometry. Consistent with the
meta-feature analysis, the prototype-distance feature family itself
received zero coefficients in the selected elastic-net
meta-classifier.

Predictive uncertainty was also examined for the standalone
EcoFuseNet-V2 branch using MC dropout. Incorrect predictions
($n=146$) exhibited a higher mean predictive entropy of 1.7940 than
correct predictions ($n=224$), for which the mean entropy was 1.5115.
The directional difference was statistically supported by the
one-sided Mann--Whitney $U$ test
($p=1.18\times10^{-10}$), with an effect size of
Cohen's $d=0.716$ under the implementation defined in
Section~\ref{subsubsec:neural-diagnostic-methods}. These results
indicate that higher EcoFuseNet-V2 predictive entropy was associated
with incorrect classifications, although the entropy distributions
were not completely separated. This uncertainty analysis applies only
to EcoFuseNet-V2 and should not be interpreted as an uncertainty
estimate for Calibrated EcoTreeFuseNet-Plus.

\subsection{Repeated-Split Stability and Computational Cost}
\label{subsec:stability-runtime-results}

The complete calibrated pipeline was repeated using seeds 42, 101,
202, 303, and 404. The seed-specific results are reported in
Table~\ref{tab:results-repeated-seed-stability}.

\begin{table}[H]
\centering
\caption{Repeated-seed and split stability of Calibrated
EcoTreeFuseNet-Plus. Variability is reported using the sample
standard deviation.}
\label{tab:results-repeated-seed-stability}
\small
\resizebox{\linewidth}{!}{%
\begin{tabular}{lrrrrrrr}
\toprule
\textbf{Seed} &
\textbf{Accuracy} &
\textbf{Macro F1} &
\textbf{Bal. Acc.} &
\textbf{MCC} &
\textbf{ECE} &
\textbf{Brier} &
\textbf{NLL} \\
\midrule
42
& 0.8000 & 0.7768 & 0.7903 & 0.7903 & 0.0651 & 0.3108 & 0.8179 \\

101
& 0.7919 & 0.7840 & 0.7894 & 0.7815 & 0.0248 & 0.3029 & 0.6302 \\

202
& 0.7865 & 0.7596 & 0.7594 & 0.7756 & 0.0568 & 0.3119 & 0.6499 \\

303
& 0.7811 & 0.7781 & 0.7692 & 0.7704 & 0.0469 & 0.3233 & 0.6891 \\

404
& 0.7865 & 0.7602 & 0.7593 & 0.7766 & 0.0516 & 0.3239 & 0.7063 \\
\midrule
Mean $\pm$ sample SD
& 0.7892 $\pm$ 0.0072
& 0.7717 $\pm$ 0.0112
& 0.7735 $\pm$ 0.0154
& 0.7789 $\pm$ 0.0075
& 0.0491 $\pm$ 0.0151
& 0.3146 $\pm$ 0.0090
& 0.6987 $\pm$ 0.0732 \\
\bottomrule
\end{tabular}}
\end{table}

Across the five runs, accuracy was $0.7892\pm0.0072$, macro F1 was
$0.7717\pm0.0112$, balanced accuracy was $0.7735\pm0.0154$, and
MCC was $0.7789\pm0.0075$.

The probability-reliability metrics were also comparatively stable.
Mean ECE was $0.0491\pm0.0151$, the Brier score was
$0.3146\pm0.0090$, and NLL was $0.6987\pm0.0732$. All variability
estimates use the sample standard deviation.

The five repeated runs of the proposed calibrated pipeline required
72.63~min in total, corresponding to $14.53\pm3.20$~min per run.
These timings exclude the complete baseline, ablation, and diagnostic
benchmark suite reported for seed 42. Because each seed generated a
separate outer split, the results represent repeated-split stability
rather than an ensemble of identical test observations.
\clearpage

\section{Discussion}
\label{sec:discussion}

\subsection{Principal Findings}
\label{subsec:discussion-principal-findings}

This study evaluated whether calibrated tree--neural probability
fusion could provide reliable fine-grained vegetation-community
classification under a limited-sample, multiclass ecological setting.
On the primary seed-42 test partition, Calibrated
EcoTreeFuseNet-Plus achieved an accuracy of 0.8000, a macro F1-score
of 0.7768, a balanced accuracy of 0.7903, and an MCC of 0.7903.
Across five repeated splits, the macro F1-score was
$0.7717\pm0.0112$, indicating limited sensitivity of the principal
finding to the selected random partition.

The discrimination improvement over the strongest standalone baseline
was modest. ExtraTrees achieved a macro F1-score of 0.7712, producing
an observed difference of 0.0055 in favour of the proposed model.
However, the paired-bootstrap confidence interval crossed zero, and
neither the paired-bootstrap test nor McNemar's test identified a
significant difference from ExtraTrees. The proposed framework should
therefore not be interpreted as demonstrating universal predictive
superiority over strong tree ensembles. Its principal contribution is
instead the combination of competitive discrimination, substantially
improved probability calibration, repeated-split stability, and a
transparent assessment of the information sources retained by the
fusion model.

This interpretation is consistent with evidence that tree ensembles
remain highly competitive for heterogeneous tabular-learning problems,
particularly when sample sizes are limited and the predictor dimension
is modest \cite{grinsztajn2022why,shwartzziv2022tabular}. The results
also support validation-based stacking rather than assuming that the
inclusion of additional model components will necessarily improve
generalization \cite{wolpert1992stacked,vanderlaan2007super}.

\subsection{Discrimination--Calibration Trade-off}
\label{subsec:discussion-calibration}

The clearest improvement provided by the proposed framework concerned
probability reliability. Temperature scaling reduced ECE from 0.3866
to 0.0651, the Brier score from 0.4839 to 0.3108, and NLL from 1.2926
to 0.8179. Because calibration did not alter the predicted class
labels, accuracy, macro F1, balanced accuracy, and MCC remained
unchanged. The improvement therefore reflects a correction of model
confidence rather than a change in class decisions.

The selected temperature was $T=0.50$, indicating that the raw
EcoTreeFuseNet-Plus probabilities were predominantly underconfident
and required sharpening. The reliability diagrams in
Figure~\ref{fig:results-reliability} show that empirical accuracy
generally exceeded mean confidence before calibration, whereas the
calibrated probabilities were substantially closer to the ideal
reliability relationship.

The simultaneous reductions in ECE, Brier score, and NLL are
important because these measures capture complementary aspects of
probabilistic performance. ECE summarizes the agreement between
confidence and empirical accuracy within probability bins, whereas
the Brier score and NLL assess the complete predictive distribution
and penalize inaccurate probability assignments
\cite{gneiting2007strictly}.

Reliable probabilities are particularly important in ecological
classification because model confidence may be used to prioritize
field verification, identify ambiguous observations, or support
subsequent environmental decision-making. A classifier with
competitive discrimination but poorly scaled confidence can create
an unjustified impression of certainty. Post-hoc calibration provides
a comparatively simple mechanism for reducing this risk without
retraining the underlying classifier
\cite{niculescu2005predicting,guo2017calibration}. In the present
study, the calibrated model therefore provided a more favourable
discrimination--calibration balance than the raw fusion output.

\subsection{Contribution of the Tree and Neural Branches}
\label{subsec:discussion-meta-features}

The selected elastic-net meta-classifier provides direct evidence
about how the final predictive distribution was formed.
Tree-probability features accounted for 93.70\% of the normalized
absolute coefficient contribution, whereas EcoFuseNet-V2
probabilities accounted for 6.19\%. The original predictors
contributed 0.11\%, while the neural embeddings, gate outputs, and
prototype-distance features received zero coefficients.

These findings indicate that the final model operated primarily as a
probability-level ensemble of tree learners rather than as a
representation-level neural fusion model. The neural branch was not
entirely excluded because its class-probability outputs supplied a
small but retained contribution. However, the learned embeddings,
gates, and prototype distances did not provide sufficient additional
validation benefit to survive elastic-net regularization.
Consequently, the final performance should not be attributed to the
latent neural representations.

The dominance of tree probabilities is plausible for the present
problem. The predictor matrix contained eight tabular variables and
1,833 observations distributed across 29 classes. Under this regime,
tree ensembles can represent nonlinear thresholds and feature
interactions without requiring the larger sample sizes generally
needed to stabilize high-dimensional neural representations. The
result also demonstrates the value of sparse meta-learning: rather
than forcing every candidate feature family into the final model, the
validation-selected elastic-net solution suppressed components that
did not improve generalization.

The contribution percentages shown in
Figure~\ref{fig:results-meta-feature} represent normalized absolute
coefficient magnitudes rather than causal feature importance.
Correlated predictors may share or redistribute coefficient magnitude,
and a zero coefficient indicates exclusion by the selected sparse
model rather than proof that the corresponding representation contains
no information \cite{zou2005regularization}. The figure should
therefore be interpreted as evidence about the fitted fusion mechanism,
not as a universal ranking of ecological information content.

\subsection{Architectural Ablation and Multisource Complementarity}
\label{subsec:discussion-ablation}

\begin{figure}[H]
    \centering
    \includegraphics[width=0.92\linewidth]{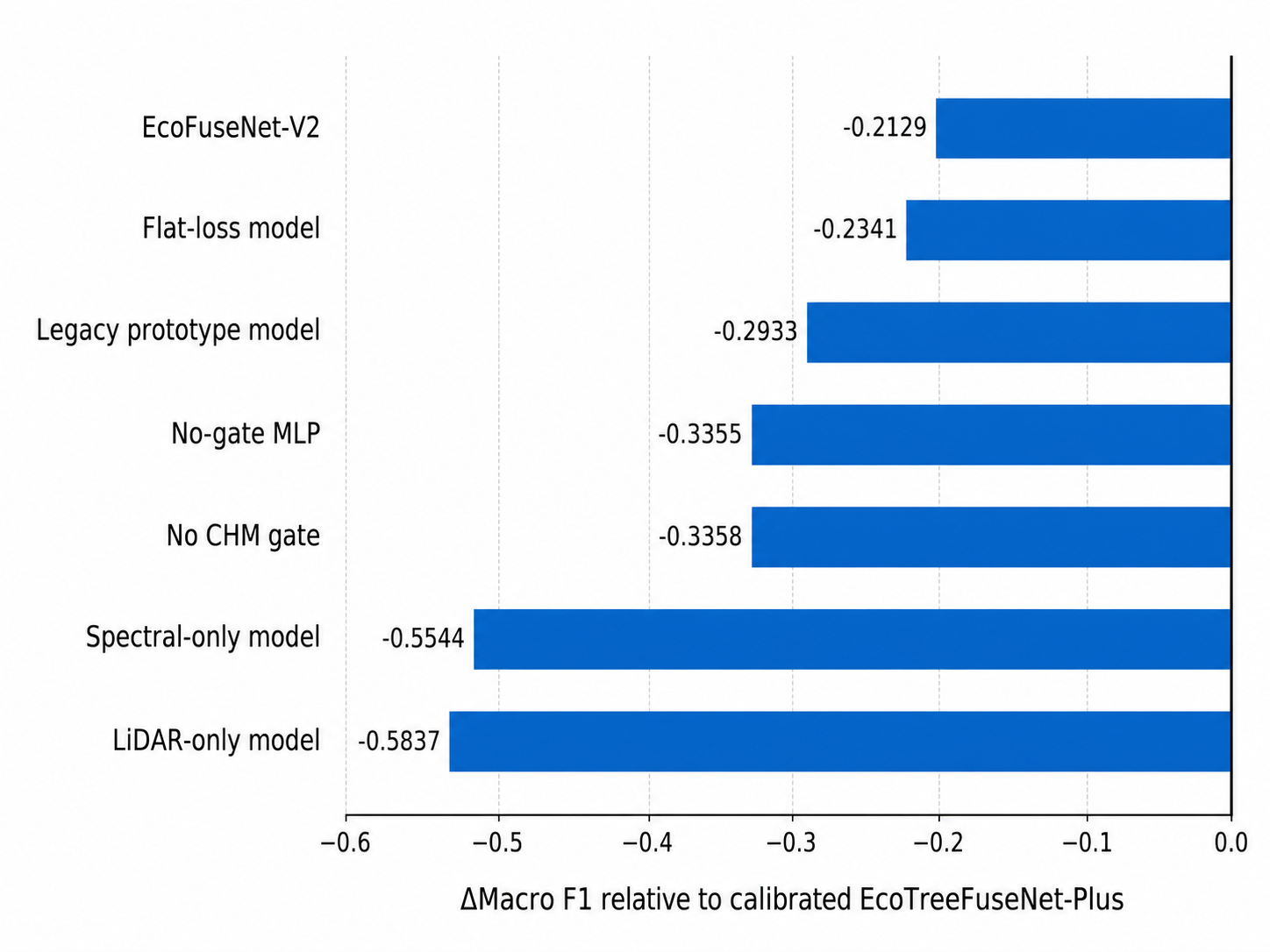}
    \caption{Change in macro F1 relative to Calibrated
    EcoTreeFuseNet-Plus for the principal architectural and modality
    ablations. All evaluated variants produced lower macro F1, with
    the largest reductions observed for the LiDAR-only and
    spectral-only configurations.}
    \label{fig:discussion-ablation}
\end{figure}

The ablation analysis examined whether simplified neural and
modality-restricted configurations could match the complete calibrated
hybrid pipeline. Figure~\ref{fig:discussion-ablation} presents the
change in macro F1 relative to Calibrated EcoTreeFuseNet-Plus. All
evaluated variants produced lower macro F1, with reductions ranging
from 0.2129 for standalone EcoFuseNet-V2 to 0.5837 for the
LiDAR-only configuration.

The largest reductions were observed for the LiDAR-only and
spectral-only models, with $\Delta$Macro F1 values of $-0.5837$ and
$-0.5544$, respectively. These results indicate that neither
information domain was sufficient independently for the 29-class
classification task. Terrain and canopy variables characterize
topographic setting and vegetation structure, whereas NDVI and NDWI
describe spectral responses associated with vegetation greenness and
moisture. Their combined use therefore provides complementary
information consistent with the broader rationale for multisource
ecological remote sensing
\cite{kampe2010neon,lefsky2002lidar,baltrusaitis2019multimodal,
albanwan2024imagefusion}.

The flat-loss variant reduced macro F1 by 0.2341, while the legacy
prototype model reduced it by 0.2933. The No-gate MLP and
No-CHM-gate models produced similar reductions of 0.3355 and 0.3358,
respectively. These comparisons show that none of the simplified
neural or modality-restricted configurations matched the complete
calibrated hybrid pipeline. Because each variant differs from the
final model in multiple respects, the reported reductions should not
be interpreted as isolated causal effects of individual architectural
components.

The ablation results must also be distinguished from ecological
interpretation of the learned gate. The relationships between
class-level gate values and CHM statistics were weak and
non-significant, and the gate outputs received zero coefficient
contribution in the selected elastic-net meta-classifier. Although the
gate-based variants performed better than their simplified
counterparts, the present results do not establish the mechanism
responsible for this difference. The non-significant CHM associations
and zero meta-level gate contribution prevent a strong ecological
interpretation of the learned gate.

\subsection{Class-Level Behaviour and Error Structure}
\label{subsec:discussion-class-behaviour}

The class-wise results demonstrate that predictive performance was not
uniform across the 29 categories. IrrigLawnPasture, Water, and
Wyethia obtained high F1-scores, whereas DrySubalpTill,
SandstoneOutcrop, SandstoneTalus, GraniteTalus, and Paved were more
difficult to classify. Some low-support classes, including BareSoil
and Frasera, achieved high F1-scores, but these estimates were based
on only four test observations and should therefore be interpreted
cautiously.

The confusion matrix in Figure~\ref{fig:results-confusion} shows that
the principal errors were concentrated among ecologically or
physically similar categories. DrySubalpTill and Festuca formed the
largest bidirectional confusion pair, followed by AlpineSandstone and
AlpineShale, and Artemisia and DrySubalpTill. Additional confusion
occurred between Paved and StreamSandGravel and between
GraniteOutcrop and SandstoneOutcrop.

These patterns are consistent with the fine-grained nature of the
target taxonomy. Several classes may share similar vegetation cover
or spectral responses while differing primarily in substrate,
topographic setting, moisture regime, or community composition.
Conversely, within-class variation in canopy condition, illumination,
and local terrain can broaden the observed predictor distributions.
Under such conditions, overall accuracy can obscure weak performance
in low-support categories, supporting the joint use of macro F1,
balanced accuracy, MCC, and class-wise evaluation
\cite{foody2002status,stehman2019key}.

The observed error structure further suggests that future gains may
depend more on additional discriminative ecological information than
on increased classifier complexity alone. More detailed vegetation
structure, seasonal spectral observations, substrate descriptors, or
additional reference samples may help separate classes that currently
occupy overlapping regions of the eight-variable predictor space.

\subsection{Uncertainty Signal in EcoFuseNet-V2}
\label{subsec:discussion-uncertainty}

MC dropout uses repeated stochastic forward passes to approximate
predictive uncertainty in neural networks \cite{gal2016dropout}. In
the present analysis, incorrect EcoFuseNet-V2 predictions had a mean
predictive entropy of 1.7940, compared with 1.5115 for correct
predictions. The one-sided Mann--Whitney test evaluated the
directional hypothesis that incorrect predictions would exhibit
higher entropy than correct predictions. The difference was
statistically supported ($p=1.18\times10^{-10}$) and had a
standardized effect size of $d=0.716$.

\begin{figure}[H]
    \centering
    \includegraphics[width=0.90\linewidth]{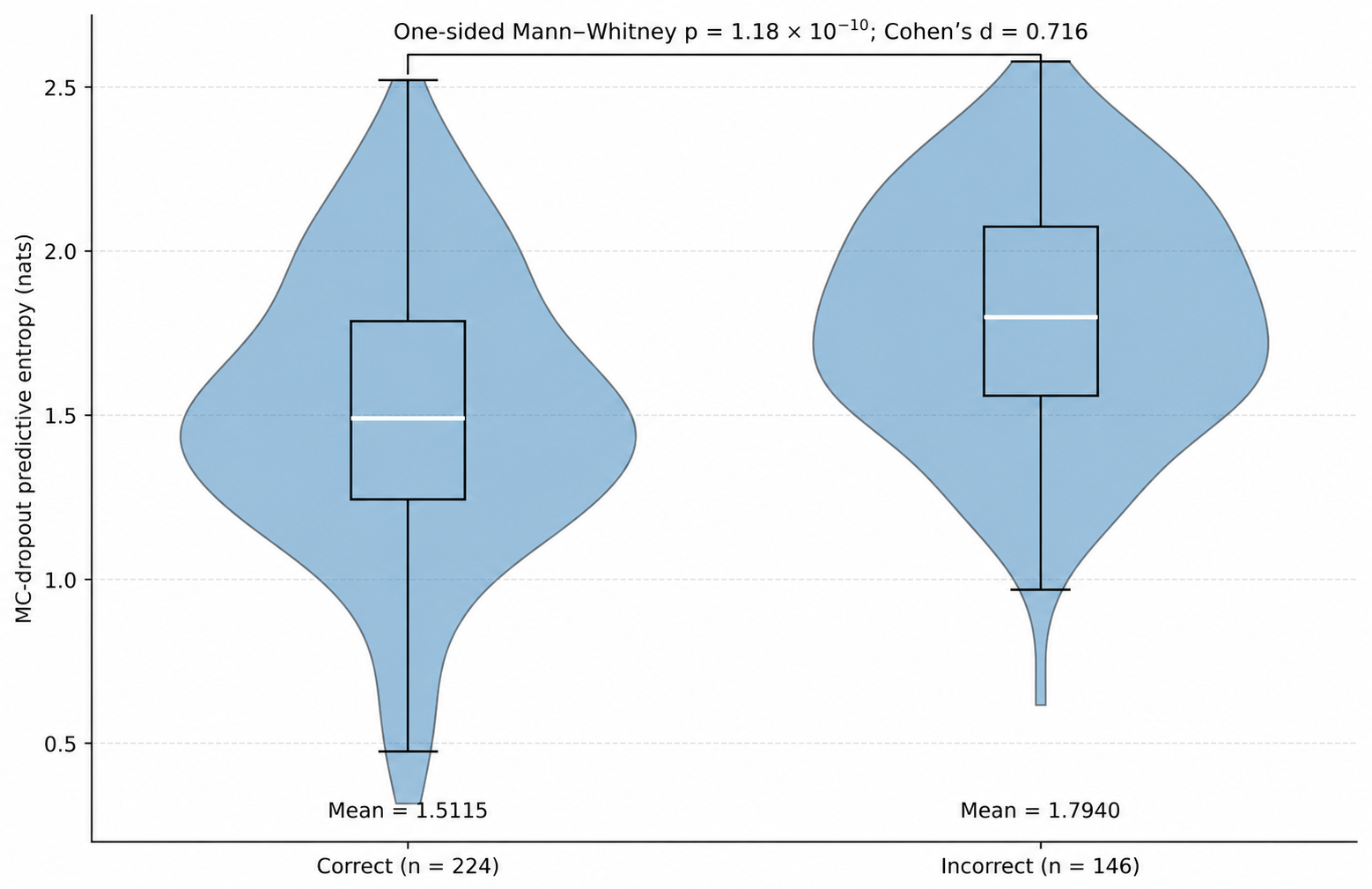}
    \caption{Distribution of MC-dropout predictive entropy for
    correct and incorrect EcoFuseNet-V2 classifications. Incorrect
    predictions exhibited higher mean entropy than correct
    predictions (1.7940 versus 1.5115; one-sided Mann--Whitney
    $p=1.18\times10^{-10}$; Cohen's $d=0.716$). This diagnostic
    applies to EcoFuseNet-V2 and not to the calibrated final
    probability-fusion model.}
    \label{fig:discussion-entropy}
\end{figure}

The distributions shown in Figure~\ref{fig:discussion-entropy}
demonstrate a clear shift toward higher entropy for incorrect
predictions, although the groups still overlap. Predictive entropy
can therefore support the relative identification of difficult neural
predictions, but it should not be treated as a deterministic error
detector. Some correct predictions remain uncertain, while some
incorrect predictions receive comparatively low entropy.

This result applies specifically to EcoFuseNet-V2. It does not
directly establish uncertainty discrimination for Calibrated
EcoTreeFuseNet-Plus because the final framework combines multiple
tree and neural probability sources and subsequently applies
temperature scaling. A complete uncertainty analysis of the final
fusion model would require uncertainty measures calculated from its
calibrated predictive distribution, ideally under both in-domain and
spatially shifted evaluation conditions.

\subsection{Repeated-Split Stability and Computational Considerations}
\label{subsec:discussion-stability}

The five repeated splits produced comparatively small variation in
the principal evaluation metrics. Macro F1 was
$0.7717\pm0.0112$, balanced accuracy was
$0.7735\pm0.0154$, MCC was $0.7789\pm0.0075$, and ECE was
$0.0491\pm0.0151$. The consistency of these values indicates that
the principal conclusions were not driven exclusively by the seed-42
partition.

The repeated results should not be interpreted as an ensemble because
each seed generated a different training, validation, and test
partition. They instead quantify sensitivity to repeated data
splitting. The observed stability strengthens confidence in the
reported discrimination--calibration trade-off, but it does not
establish transfer to a different watershed, acquisition year, sensor
configuration, or ecological region.

The complete seed-42 benchmark suite required 20.56~min on a CPU,
whereas the five repeated runs of the proposed calibrated pipeline
required 72.63~min in total. This computational requirement is
moderate for model development and repeated validation. However, the
fusion pipeline is more complex than a standalone ExtraTrees model,
while its macro-F1 improvement over ExtraTrees was small and
non-significant. The additional complexity is therefore most
defensible when calibrated probabilities, multi-model robustness, or
uncertainty-aware analysis are required. For applications concerned
only with hard class labels, a strong tree ensemble may provide a
simpler and competitive alternative.

\clearpage

\section{Conclusions}
\label{sec:conclusions}

This study developed Calibrated EcoTreeFuseNet-Plus for fine-grained vegetation-community classification by integrating terrain, canopy-structure, and spectral predictors through a calibrated tree--neural fusion strategy. The framework achieved an accuracy of 0.8000, a macro F1-score of 0.7768, a balanced accuracy of 0.7903, and an MCC of 0.7903, while repeated-split evaluation showed stable overall performance. Its principal advantage was improved probability reliability rather than a large discrimination gain over ExtraTrees, as temperature scaling reduced ECE from 0.3866 to 0.0651, the Brier score from 0.4839 to 0.3108, and NLL from 1.2926 to 0.8179 without changing the predicted classes. The final meta-classifier was driven mainly by tree probabilities, with a smaller complementary contribution from EcoFuseNet-V2 probabilities, and the neural uncertainty analysis showed that incorrect predictions generally had higher predictive entropy than correct predictions. However, the evaluation was limited to one watershed and one airborne survey, several classes had low sample support, the canopy-height variable was highly sparse, and the out-of-fold procedure was leakage-aware but not fully nested. In addition, the reported entropy analysis was restricted to EcoFuseNet-V2 and did not directly quantify uncertainty in the complete calibrated fusion model. Future work should therefore evaluate the framework using independent sites, spatially blocked and fully nested validation, additional temporal observations, and richer vegetation-structure information. It should also extend uncertainty estimation and calibration analysis to the complete fusion model under spatial and temporal distribution shifts. In practical applications, the calibrated predictions could support vegetation-community mapping, ecological inventory, restoration planning, habitat monitoring, and targeted field verification by identifying locations requiring additional observation. Overall, the study demonstrates that carefully calibrated tree--neural fusion can provide a reliable and transparent decision-support framework for fine-grained ecological classification while clearly defining the validation required before broader operational deployment.
\bibliographystyle{cas-model2-names}

\bibliography{cas-refs}


\end{document}